\documentclass[journal]{IEEEtran}

\usepackage{amsmath,amssymb}
\usepackage{graphicx}
\usepackage{booktabs}
\usepackage{array}
\usepackage{algorithm}
\usepackage{algpseudocode}
\usepackage{xcolor}
\usepackage{cite}
\usepackage[hidelinks]{hyperref}

\graphicspath{{figures/}{figures/plots/}{figures/diagrams/}{images/}{./}}

\begin{document}

\title{The Multipath Blind Spot: $K$-Agnostic Robust Calibration for
Sparse-Anchor Metric Depth from Frozen Foundations}

\author{Sohag Roy,~Rajesh Misra,~Swami Shastravidyananda,~and Tamal Maharaj
	\thanks{The authors are with the Department of Computer Science,
		Ramakrishna Mission Vivekananda Educational and Research Institute (RKMVERI),
		Belur, Howrah, West Bengal 711202, India.
		E-mail: roysohag95@gmail.com; rajesh2025.misra@gmail.com;
		shastravidyananda@gm.rkmvu.ac.in; tamal@gm.rkmvu.ac.in.}%
}

\markboth{Under review at IEEE Transactions on Pattern Analysis and Machine Intelligence}%
{Roy \MakeLowercase{\textit{et al.}}: The Multipath Blind Spot}

\maketitle

\begin{abstract}
Monocular depth foundations predict domain-general relative depth but lack
absolute scale; a handful of sparse metric anchors from a range sensor can
calibrate them to metric depth, an attractive alternative to
metric-supervised training. Existing sparse-anchor calibration methods,
however, assume the anchors are clean, whereas real sensors produce outliers
that are present with the wrong value---time-of-flight multipath, mixed
pixels---not merely missing. We show that the established residual-on-CFA
calibration recipe collapses under such outliers, and that the strongest
publicly deployed method, VI-Depth, has a structural multipath blind spot:
robust to missing anchors, it falls behind an unprotected baseline on three
of four datasets when anchors are present but wrong. We propose
Multipath-Robust Anchor Calibration (MRAC), a parameter-free, inference-time
wrapper that gates anchors by foundation consistency---a Theil--Sen fit and a
median-absolute-deviation test against the foundation's own relative-depth
ordering---before a single call to the calibration head. MRAC adds no learned
parameters, runs its selection in $\approx 50\,\mu$s on CPU, and serves anchor
budgets $K \in [5,200]$ from one checkpoint. On a $320$-cell benchmark with a
same-backbone, same-architecture control, MRAC strictly wins $84\%$ of
same-backbone cells across all four outlier families and, against VI-Depth,
wins all twelve corrupted multipath cells and all sixteen KITTI cells,
reducing KITTI multipath AbsRel by $3.2\times$ ($0.489$ to $0.151$) at zero
retraining.
\end{abstract}


% =====================================================================
\section{Introduction}
\label{sec:intro}

\IEEEPARstart{M}{onocular} depth foundations now predict strong,
domain-general relative depth from a single image \cite{Ranftl2022,
Ranftl2021, Yang2024b}. Their output is affine-invariant: it encodes scene
structure but carries no absolute scale. A small set of sparse metric
anchors---a few hundred or even a few dozen points from a time-of-flight
sensor, LiDAR, or structured-light module---is enough to recover that scale
and turn a relative-depth foundation into a metric one. This
frozen-foundation-plus-sparse-anchor paradigm is attractive precisely because it is light: it
inherits the foundation's cross-domain prior, needs no metric-supervised
retraining, and runs at interactive rates. VI-Depth \cite{Wofk2023}, the
sparse-LiDAR rescaling of Marsal et al. \cite{Marsal2025}, and a growing line
of recent methods \cite{Lin2025, Wang2025b, Zheng2026} all follow it.

These methods share an unstated assumption: that the anchors are clean. Real
range sensors violate it. Time-of-flight multipath off glass and specular
surfaces returns depths biased toward the sensor; LiDAR saturates at maximum
range and reports a constant sentinel value; beam footprints that straddle a
depth discontinuity return a blend of foreground and background. Most of
these corruptions are \emph{present with the wrong value}, not missing---and
a present-but-wrong anchor is categorically harder than a missing one,
because a calibrator cannot simply skip a measurement that looks valid. Yet
no prior work systematically measures how sparse-anchor calibration behaves
under sensor outliers.

\begin{figure*}[!t]
\centering
\includegraphics[width=\textwidth]{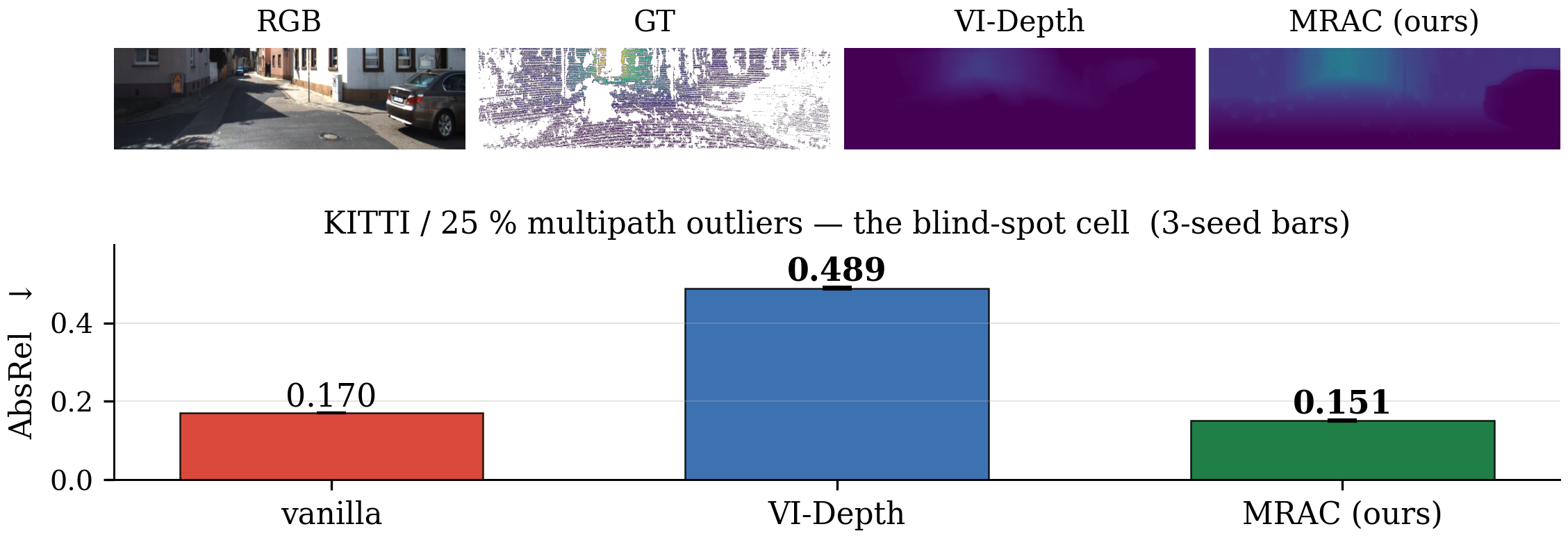}
\caption{Sparse-anchor calibration under multipath outliers. The strongest
deployed method, VI-Depth, is robust to missing anchors but blind to
present-but-wrong-value multipath anchors, inflating KITTI AbsRel to $0.489$
at $25\%$ corruption. MRAC, a parameter-free inference-time wrapper, gates
the foundation-inconsistent anchors and reduces this to $0.151$, a
$3.2\times$ improvement at zero retraining.}
\label{fig:teaser}
\end{figure*}

We measure it, and the result is surprising. VI-Depth, the strongest
publicly deployed method of this kind, has a structural blind spot on
multipath outliers. It is robust to anchor dropout---the corruption its
training implicitly simulates---but on present-with-wrong-value multipath
anchors its accuracy degrades sharply, falling behind even an unprotected
baseline on three of four datasets. On KITTI at $25\%$ multipath corruption
its AbsRel is $0.489$, a $3.2\times$ error inflation over the method we
propose (Fig.~\ref{fig:teaser}). The failure is not a tuning artifact: it
persists at VI-Depth's own training anchor budget, and it follows from a
missing mechanism. The pipeline has no step that tests whether a present
anchor's value is consistent with the scene geometry the foundation already
encodes.

We propose Multipath-Robust Anchor Calibration (MRAC), a parameter-free,
inference-time wrapper that supplies exactly that test. The frozen
foundation's relative-depth field is an independent witness on scene
geometry: legitimate anchors obey a single affine relationship with it, and
outliers---whether multipath, dropout, or mixed-pixel---violate it. MRAC fits
this relationship robustly with a Theil--Sen estimator, gates anchors by a
median-absolute-deviation test, and passes only the consistent anchors to a
single call of the established residual-on-CFA calibration head. It adds no
learned parameters and no $K$-dependent ones, calls the head exactly once,
runs its selection in $\approx 50\,\mu$s on CPU, and serves
anchor budgets from $K{=}5$ to $K{=}200$ from a single checkpoint---in
contrast to the per-budget heads the deployed alternative requires.

This paper makes four contributions.
\begin{itemize}
\item We identify and explain a \emph{structural multipath blind spot} in
the strongest publicly deployed sparse-anchor calibration method, showing it
is robust to missing anchors but fails on present-with-wrong-value ones, and
that the failure persists at its own training anchor budget
(Secs.~\ref{sec:robustness},~\ref{sec:discussion}).
\item We propose \emph{MRAC}, a parameter-free, single-forward-pass,
$K$-agnostic inference-time wrapper that closes this gap by gating anchors on
foundation consistency, at $\approx 50\,\mu$s CPU overhead and no added
parameters (Secs.~\ref{sec:method},~\ref{sec:cost}).
\item We construct the \emph{first systematic anchor-outlier robustness
benchmark} for this setting, to our knowledge: $320$ cells spanning four
datasets, four sensor-grounded outlier families, four corruption fractions,
and five methods, with a same-backbone, same-architecture control that
isolates the method from the backbone (Sec.~\ref{sec:setup}).
\item We show MRAC strictly wins $84\%$ of cells against its same-backbone
controls across all four outlier families, and against VI-Depth wins all
twelve corrupted multipath cells and all sixteen KITTI cells, reducing KITTI
multipath AbsRel by $3.2\times$ at zero retraining
(Sec.~\ref{sec:robustness}).
\end{itemize}

Section~\ref{sec:related} situates these contributions, Section~\ref{sec:method}
defines MRAC, and Sections~\ref{sec:setup}--\ref{sec:cost} present the
benchmark, the headline robustness result, the substrate and diagnostic
studies, and the inference cost; Section~\ref{sec:discussion} discusses
limitations and Section~\ref{sec:conclusion} concludes.

% =====================================================================
\section{Related Work}
\label{sec:related}

\subsection{Monocular Depth Foundations}
\label{sec:rel:foundations}
Affine-invariant monocular depth estimation has matured into general-purpose
foundations. MiDaS established robust relative-depth prediction by training
across mixed datasets with a scale-and-shift-invariant loss
\cite{Ranftl2022}; DPT replaced the convolutional backbone with a vision
transformer \cite{Ranftl2021}; and Depth Anything scaled training to large
unlabeled corpora \cite{Yang2024a}, with Depth Anything V2 improving fine
detail and robustness \cite{Yang2024b}. Diffusion-based estimation reaches
comparable affine-invariant accuracy \cite{Ke2024}. The appeal of these
models is that a single frozen network supplies a strong geometric prior
across domains without per-scene or per-sensor training. Other foundations
target metric scale directly---monocular geometry with metric output
\cite{Wang2025}, fast metric depth \cite{Bochkovskii2024}, and zero-shot
metric prediction \cite{Yin2023}---though typically at the cost of generality
or with additional metric supervision. The relative-depth models output a
field that encodes scene structure but no absolute scale; recovering metric
depth from them requires an external cue, which is the problem this paper
addresses. We keep the foundation frozen throughout and never fine-tune it.

\subsection{Metric Depth from Sparse Cues}
\label{sec:rel:sparsecue}
One route to metric depth is full supervision, regressing metric depth
directly: from the multi-scale network of Eigen et al. \cite{Eigen2014}
through ordinal regression \cite{Fu2018}, local planar guidance
\cite{Lee2019}, adaptive bins \cite{Bhat2021}, neural-window CRFs
\cite{Yuan2022}, and the relative-plus-metric combination of ZoeDepth
\cite{Bhat2023}. These require metric-supervised training and do not exploit
a frozen foundation at inference.

The route we follow instead calibrates a frozen relative-depth model with a
handful of sparse metric anchors---a lightweight alternative that inherits
the foundation's cross-domain prior. VI-Depth aligns monocular depth to
sparse visual-inertial points by a global scale-and-shift fit followed by a
learned dense alignment head, and is the most widely deployed method of this
kind \cite{Wofk2023}. Marsal et al. rescale Depth Anything to a 2D-LiDAR's
sparse points by linear regression, using RANSAC for robustness
\cite{Marsal2025}. More recent work conditions on denser or more general
cues: Prompt Depth Anything fuses a low-resolution LiDAR depth as a decoder
prompt for 4K metric depth \cite{Lin2025}, Prior Depth Anything incorporates
arbitrary depth priors \cite{Wang2025b}, and SLIM injects sparse LiDAR into a
geometry foundation for long-range driving \cite{Zheng2026}. The closed-form
fit plus learned residual that we adopt as a calibration substrate is
established by VI-Depth and Marsal et al.; our contribution is orthogonal to
it. Critically, none of these methods addresses anchor outliers. They assume
the sparse cues are clean---Marsal's RANSAC stage being the only explicit
robustness, and one we show fails on correlated dropout
(Sec.~\ref{sec:robustness})---whereas real range sensors produce multipath,
dropout, and mixed-pixel corruption.

\subsection{Robust Regression in Vision}
\label{sec:rel:robust}
Robust estimation supplies the fitting tools for corrupted data. The
Theil--Sen estimator takes the median of pairwise slopes and attains a
breakdown point of $\approx 29\%$ \cite{Theil1950,Sen1968}; least median of
squares \cite{Rousseeuw1984} and M-estimators \cite{Huber1964} provide
alternative robust criteria. In vision, RANSAC \cite{Fischler1981} and its
maximum-likelihood variant MLESAC \cite{Torr2000} are the standard consensus
estimators for geometric fitting \cite{Hartley2004}. Consensus sampling is
known to degrade when outliers are structured or mutually consistent rather
than independent, because a coherent outlier set can itself form a
high-consensus model. The anchor-outlier setting contains exactly this case:
dropout outliers share one constant depth and are therefore correlated. We
adapt the median-based Theil--Sen fit with a MAD inlier gate and use the
frozen foundation's relative depth as the consistency reference, which is
what lets the fit reject correlated outliers that a consensus step would
accept (Secs.~\ref{sec:method:mrac},~\ref{sec:robustness}).

\subsection{Sparse Depth Completion}
\label{sec:rel:completion}
The adjacent field of depth completion predicts a dense depth map from a
sparse depth input and an image, from early image-guided CNNs
\cite{Ma2018,Ma2019} through spatial-propagation networks
\cite{Cheng2018,Park2020}, guided convolution \cite{Tang2021,Hu2021},
transformer hybrids \cite{Zhang2023}, bilateral propagation \cite{Tang2024},
and unsupervised calibrated backprojection \cite{Wong2021}. Completion
differs from our problem in two ways. It learns the full dense prediction,
which requires heavy training specific to a sensor and a sparsity pattern,
whereas we calibrate a frozen foundation with a single small head that is
agnostic to the anchor budget. And, like the sparse-cue calibration methods
above, completion networks take the sparse input at face value; outlier
robustness is outside their scope. Our method is complementary: it makes the
calibration of a frozen foundation robust to the corrupted measurements that
both calibration and completion otherwise trust.

% =====================================================================
\section{Method}
\label{sec:method}

\subsection{Problem Setup}
\label{sec:method:setup}
\begin{figure}[t]
\centering
\includegraphics[width=\columnwidth]{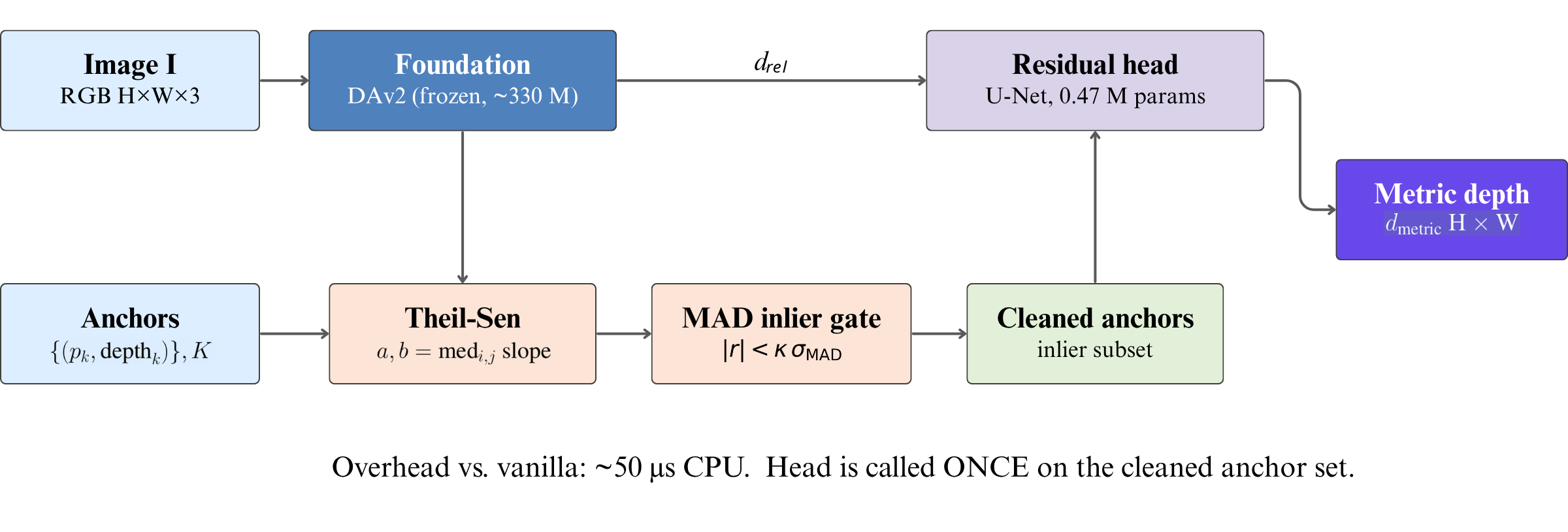}
\caption{MRAC pipeline. A frozen foundation $F$ (DAv2, ${\sim}330$\,M
	parameters) predicts a relative-depth field $d_{\mathrm{rel}}$; the $K$ sparse
	anchors are tested for consistency with that field by a Theil--Sen slope fit
	and a MAD inlier gate ($|r_k| \le \kappa\,\sigma_{\mathrm{MAD}}$); only the
	surviving cleaned anchors are passed to a single call of the residual-on-CFA
	head $R_\theta$ ($0.47$\,M parameters), which produces the metric depth map.
	The robust selection adds no parameters and no second forward pass
	(${\sim}50\,\mu$s CPU overhead).}
\label{fig:pipeline}
\end{figure}

We recover a dense metric depth map from a single RGB image $I$ and a
small set of sparse depth measurements. A frozen monocular depth
foundation $F$ maps $I$ to a normalized relative depth field
$d_{\mathrm{rel}} = F(I) \in [0,1]^{H\times W}$, which encodes the scene's
relative geometry but carries no absolute scale. The metric cue is a set
of $K$ sparse anchors $\{(p_k, z_k)\}_{k=1}^{K}$, where $p_k$ is a pixel
coordinate and $z_k$ is the depth reported by a range sensor (ToF, LiDAR,
or structured light) at $p_k$. The objective is the metric depth map
$D \in \mathbb{R}^{H\times W}$. Fig.~\ref{fig:pipeline} overviews the
pipeline.

The foundation is frozen throughout: no gradients propagate into $F$, and
$F$ is never specialized to a sensor, scene, or anchor budget. The number
of anchors $K$ is small ($K \in [5, 200]$ in our experiments) and is a
property of the deployed sensor, not a design choice. Crucially, and in
contrast to all prior sparse-anchor calibration work, we do not assume the
$z_k$ are clean: a fraction of the anchors may be corrupted by
sensor-realistic outlier processes (Sec.~\ref{sec:method:outliers}). The
method must therefore be robust to corrupted anchors, agnostic to $K$, and
add negligible cost over the frozen forward pass.

\subsection{Closed-Form Anchor Fit (CFA) Substrate}
\label{sec:method:cfa}
The simplest metric calibration aligns $d_{\mathrm{rel}}$ to the anchors by
a global affine transform. Writing $x_k = d_{\mathrm{rel}}(p_k)$, the
closed-form anchor fit (CFA) solves the least-squares problem
\begin{equation}
\label{eq:cfa}
(a_{\mathrm{cfa}}, b_{\mathrm{cfa}})
= \operatorname*{arg\,min}_{a,b}\;
\sum_{k=1}^{K}\big(a\,x_k + b - z_k\big)^2 ,
\end{equation}
which admits the closed-form solution
$a_{\mathrm{cfa}} = \operatorname{Cov}(x,z)/\operatorname{Var}(x)$ and
$b_{\mathrm{cfa}} = \bar{z} - a_{\mathrm{cfa}}\,\bar{x}$, costing a constant
number of operations per image. The metric depth map is then
$D(u) = a_{\mathrm{cfa}}\,d_{\mathrm{rel}}(u) + b_{\mathrm{cfa}}$ at every
pixel $u$.

This global-affine alignment is established prior art, not a contribution
of this work. It is the global scale-and-shift alignment stage of VI-Depth
\cite{Wofk2023}, the pseudo-metric rescaling of Marsal \emph{et al.}
\cite{Marsal2025}, and the standard scale-shift recovery used throughout
the affine-invariant depth literature \cite{Ranftl2022}. We adopt CFA
unchanged as the \emph{calibration substrate} on which both the learned
residual (Sec.~\ref{sec:method:residual}) and our robust selection
(Sec.~\ref{sec:method:mrac}) operate. The CFA estimator is the
maximum-likelihood affine fit under i.i.d.\ Gaussian anchor noise, and for
$K \ge 2$ it is exact whenever $\operatorname{Var}(x) > 0$; its weakness is
that least squares has a breakdown point of zero, so a single gross outlier
can dominate the fit. That weakness is the subject of
Sec.~\ref{sec:method:mrac}.

\subsection{Residual-on-CFA Architecture}
\label{sec:method:residual}
A purely global affine map cannot correct spatially varying
misalignment between $d_{\mathrm{rel}}$ and the true depth. The community's
response is to add a lightweight network that predicts a per-pixel
correction to the affine coefficients. A compact U-Net $R_\theta$
($0.47$\,M trainable parameters) takes the image, the relative depth, and
the anchors rendered as a sparse mask and depth channel,
\begin{equation}
\label{eq:residual_head}
\big(\Delta a, \Delta b, s\big)
= R_\theta\!\big(I,\, d_{\mathrm{rel}},\, \mathbf{m},\, \mathbf{z}\big),
\end{equation}
and emits per-pixel residual coefficients $\Delta a(u), \Delta b(u)$ and a
log-variance $s(u)$ used by an uncertainty-weighted training term
(Sec.~\ref{sec:method:training}). The metric depth is
\begin{equation}
\label{eq:residual_depth}
D(u) = \big(a_{\mathrm{cfa}} + \Delta a(u)\big)\,d_{\mathrm{rel}}(u)
     + \big(b_{\mathrm{cfa}} + \Delta b(u)\big).
\end{equation}
The residual branch is zero-initialized, so $\Delta a = \Delta b = 0$ at
the start of training and the network is \emph{exactly} CFA at
initialization. The architecture therefore degrades gracefully by
construction: in the worst case it reduces to the closed-form affine fit,
which is the $K\!\ge\!2$ maximum-likelihood estimator under a global affine
model, and it can only improve on CFA where the data support a spatially
varying correction.

We stress that this residual-on-CFA architecture is, again, prior art. It
is the learned dense alignment of VI-Depth \cite{Wofk2023} and the
refine-where-reliable stage of Marsal \emph{et al.} \cite{Marsal2025}; the
broader strategy of conditioning a small head on a frozen foundation plus
sparse metric cues is shared by Prompt Depth Anything \cite{Lin2025} at a
denser anchor regime. We treat residual-on-CFA as the substrate the field
has converged on. Our contribution is orthogonal to it: a robust selection
of \emph{which anchors the substrate is allowed to see}, introduced next.

\subsection{The Anchor-Outlier Problem}
\label{sec:method:outliers}
Real range sensors do not deliver clean anchors. We model four
sensor-grounded outlier processes, each of which replaces a fraction
$p$ of the anchor depths $z_k$ with a corrupted value.

\emph{Uniform} outliers replace $z_k$ with a value drawn uniformly across
the scene's depth range, modeling spurious returns and gross measurement
faults \cite{Tuley2005,Hebert1992}. \emph{Near
(multipath/specular)} outliers replace $z_k$ with a near-biased value,
modeling time-of-flight multipath off glass and reflective geometry and the
foreshortened returns of retro-reflective surfaces \cite{Fuchs2010,Hansard2013}. \emph{Dropout} outliers replace $z_k$ with a single constant value,
modeling sensor max-range saturation and no-return pixels reported as a fixed
sentinel depth \cite{Hansard2013,Hebert1992}. \emph{Mixed-pixel}
outliers replace $z_k$ with a depth blended between foreground and background,
modeling beam footprints that straddle depth discontinuities at object edges
\cite{Godbaz2013,Tuley2005}.

These families are not interchangeable, and the distinction drives our
design. Uniform, near, and mixed-pixel outliers are \emph{present with the
wrong value}; dropout outliers are additionally \emph{correlated}, because
every corrupted anchor takes the same constant depth. Correlation defeats
consensus methods such as RANSAC \cite{Fischler1981}: a sufficiently large
cluster of identical dropout anchors forms a self-consistent pseudo-inlier
set that a sampling-based estimator can mistake for the true model. A
robust estimator for this setting must reject outliers that are present
rather than missing, and must not be fooled by correlated outlier clusters.

\subsection{Multipath-Robust Anchor Calibration (MRAC)}
\label{sec:method:mrac}
We propose Multipath-Robust Anchor Calibration (MRAC), an inference-time
wrapper that identifies trustworthy anchors before calibration and feeds
only those anchors to the residual substrate. MRAC adds no learned
parameters and requires no retraining. Its operating principle is
\emph{foundation consistency}: the relative depth $d_{\mathrm{rel}}$ is an
independent witness on scene geometry, so legitimate anchors satisfy a
single affine relationship with $d_{\mathrm{rel}}$, whereas
outliers---however they were produced---violate it. MRAC has three stages.

\emph{Theil--Sen robust fit.} We estimate the affine relationship between
$x_k = d_{\mathrm{rel}}(p_k)$ and $z_k$ by the Theil--Sen estimator
\cite{Theil1950,Sen1968}, the median of pairwise slopes over all
distinct-$x$ anchor pairs,
\begin{equation}
\label{eq:theilsen}
\hat{a} = \operatorname*{median}_{i<j,\; x_i \neq x_j}
\frac{z_i - z_j}{x_i - x_j},
\qquad
\hat{b} = \operatorname*{median}_{k}\big(z_k - \hat{a}\,x_k\big).
\end{equation}
The Theil--Sen slope is threshold-free, scale-adaptive, and has an
asymptotic breakdown point of $1 - 1/\sqrt{2} \approx 29\%$. Critically, it
resists the correlated-dropout failure of RANSAC: the median of slopes is
dominated by the abundant pairs drawn from true anchors, and the
near-zero slopes contributed by dropout-pair samples remain a minority so
long as the inlier fraction exceeds the breakdown point.

\emph{MAD inlier gate.} Using the robust fit, we compute anchor residuals
$r_k = z_k - (\hat{a}\,x_k + \hat{b})$ and a robust scale via the median
absolute deviation,
\begin{equation}
\label{eq:mad}
\sigma_{\mathrm{MAD}} = 1.4826\,
\operatorname*{median}_{k}\big|\,r_k - \operatorname{median}_j r_j\,\big| .
\end{equation}
Because the Theil--Sen intercept~\eqref{eq:theilsen} is the median residual,
the residuals are median-centered, and we retain the inlier set
\begin{equation}
\label{eq:inliers}
\mathcal{M} = \{\, k : |r_k| \le \kappa\,\sigma_{\mathrm{MAD}} \,\},
\qquad \kappa = 2 .
\end{equation}
The multiplier $\kappa$ is the single deployment knob MRAC introduces; we
quantify its sensitivity at the end of this subsection.

\emph{Calibration on the cleaned set.} The residual substrate of
Sec.~\ref{sec:method:residual} is then evaluated using only the inlier
anchors $\mathcal{M}$. The substrate recomputes its internal CFA on
$\mathcal{M}$, so the affine base in~\eqref{eq:residual_depth} is now fit
to clean measurements, and the learned residual operates on a consistent
anchor set. This is a \emph{single} forward pass of $R_\theta$: MRAC
\emph{replaces} the anchor set passed to the vanilla head, it does not add
a second head call. Theil--Sen and the MAD gate are inexpensive scalar
operations on the $K$ anchors and run on CPU.

Foundation consistency is what unifies the four outlier families under one
filter. A multipath, dropout, or mixed-pixel anchor is inconsistent with
the $d_{\mathrm{rel}}$-implied affine relationship in exactly the same way
a uniform outlier is, and \eqref{eq:theilsen}--\eqref{eq:inliers} reject it
without any outlier-type-specific tuning and without per-dataset
thresholds. The same code handles indoor NYUv2 ($1$--$10$\,m) and outdoor
KITTI ($1$--$80$\,m). MRAC introduces no $K$-dependent parameters; the
robust fit and gate are defined for any $K \ge 2$, and the shared residual
head is the same across the full anchor range
(Secs.~\ref{sec:substrate},~\ref{sec:cost}). Algorithm~\ref{alg:mrac}
summarizes the procedure.

\begin{algorithm}[t]
\caption{Multipath-Robust Anchor Calibration (inference)}
\label{alg:mrac}
\begin{algorithmic}[1]
\Require image $I$; frozen foundation $F$; anchors $\{(p_k,z_k)\}_{k=1}^{K}$;
         residual head $R_\theta$; MAD multiplier $\kappa$
\Ensure metric depth map $D$
\State $d_{\mathrm{rel}} \gets F(I)$
\Comment{frozen forward pass}
\State $x_k \gets d_{\mathrm{rel}}(p_k)$ for $k = 1,\dots,K$
\State $\hat{a} \gets \operatorname{median}\{(z_i\!-\!z_j)/(x_i\!-\!x_j) : i<j,\, x_i\!\neq\! x_j\}$
\Comment{Theil--Sen}
\State $\hat{b} \gets \operatorname{median}\{z_k - \hat{a}\,x_k\}$
\State $r_k \gets z_k - (\hat{a}\,x_k + \hat{b})$ for $k = 1,\dots,K$
\State $\sigma \gets 1.4826 \cdot \operatorname{median}|r_k - \operatorname{median}(r)|$
\State $\mathcal{M} \gets \{\, k : |r_k| \le \kappa\,\sigma \,\}$
\Comment{MAD inlier gate, $\kappa\!=\!2$}
\State $(\Delta a, \Delta b, \cdot) \gets R_\theta\!\big(I, d_{\mathrm{rel}}, \mathbf{m}_{\mathcal{M}}, \mathbf{z}_{\mathcal{M}}\big)$
\Comment{single head call on cleaned anchors; CFA recomputed on $\mathcal{M}$}
\State $D \gets (a_{\mathrm{cfa}}^{\mathcal{M}} + \Delta a)\odot d_{\mathrm{rel}} + (b_{\mathrm{cfa}}^{\mathcal{M}} + \Delta b)$
\State \Return $D$
\end{algorithmic}
\end{algorithm}

\paragraph{Sensitivity to $\kappa$.}
The gate width $\kappa$ is a single scalar; we set it to $\kappa\!=\!2$,
the conventional two-MAD threshold for robust outlier rejection. A sweep
of $\kappa \in \{1.5, 2.0, 2.5, 3.0\}$ on the eight headline cells (four
datasets $\times$ \{near, dropout\} at $25\%$) confirms that $\kappa\!=\!2$
sits in a flat region: MRAC AbsRel varies by at most $0.011$ across the
swept range on six of eight cells and by at most $0.04$ on the two hardest
(DIODE dropout, KITTI near). Per-cell numbers are tabulated in the
supplement.

\subsection{Training}
\label{sec:method:training}
MRAC is an inference-time procedure and requires no dedicated training: the
residual head $R_\theta$ is trained once, on clean anchors, with the
standard recipe. We optimize for $80$ epochs with AdamW (learning rate
$10^{-3}$, $2$ warmup epochs, cosine decay), batch size $12$, and
mixed-precision (fp16), taking roughly $10$ hours on one A100; foundation
predictions are cached so $F$ is evaluated once per image over training.
Anchors are drawn uniformly at random per minibatch from
$K \in \{5, 10, 25, 50\}$, which is what makes a single trained head serve
the full deployment range without per-$K$ checkpoints. Training uses four
loss terms---a scale-invariant log term (SILog) \cite{Eigen2014} as the
primary objective, together with confidence-weighting (CWA),
structural-scale consistency (SSC), and feature-matching (FMP) terms with weights
$1.0$, $1.0$, $0.5$, and $0.2$ respectively; their interaction with the
architecture is analyzed in Sec.~\ref{sec:diagnostics}. The residual
architecture additionally carries an auxiliary regularization term
$\mathcal{L}_{\mathrm{res}} = |\Omega|^{-1} \sum_{p\in\Omega}(|a_p - a^\star| + |b_p - b^\star|)$
that penalizes the per-pixel deviation of the residual $(a,b)$ fields from
the global CFA fit $(a^\star, b^\star)$; the loss study of
Sec.~\ref{sec:diagnostics} finds it is best removed, and the recommended
configuration omits it ($w_{\mathrm{res}}\!=\!0$). No outlier injection is
used at training time. Robustness is supplied entirely at inference by MRAC,
which is what makes the method a drop-in wrapper for any already-trained
residual-on-CFA head.

\subsection{Inference Cost}
\label{sec:method:cost}
MRAC's overhead over the vanilla pipeline is the Theil--Sen fit and the MAD
gate. The Theil--Sen slope is $O(K^2)$ in the number of anchor
pairs---about $K(K\!-\!1)/2 \approx 1.2\text{k}$ slopes at $K=50$---and the MAD gate
is $O(K)$; together they cost approximately $50\,\mu$s on CPU. No new
learned parameters are added, and the residual head is called exactly once,
on the cleaned anchor set, in place of the vanilla call. The net deployment
overhead relative to vanilla inference is therefore under $0.1$\,ms, leaving
end-to-end latency dominated by the two forward passes already present in
the baseline (Sec.~\ref{sec:cost}).

% =====================================================================
\section{Experimental Setup}
\label{sec:setup}

\subsection{Datasets}
\label{sec:setup:data}
We evaluate on four datasets spanning indoor and outdoor scenes. On
\textbf{NYUv2} \cite{Silberman2012} we use the official Eigen test split
\cite{Eigen2014} of $654$ images with colorization-filled depth (the
\texttt{depths} field), the standard BTS protocol \cite{Lee2019}; we train
and test on the same filled ground truth. On \textbf{KITTI} \cite{Geiger2012}
we use the Eigen test split of $651$ images after removing $45$ frames with
no valid ground-truth line, with the standard Garg crop \cite{Garg2016} and an $80$\,m cap. \textbf{DIODE} \cite{Vasiljevic2019}
contributes its $771$-image validation set, mixing indoor and outdoor scenes.
\textbf{SUN RGB-D} \cite{Song2015} is evaluated at stride~$5$ ($2067$
images); we disclose that the official $5050$-image test split requires the
SUNRGBD-toolbox split files, which are not shipped with the public release,
and that the stride-$5$ subset yields stable averages. We drop ETH3D: its
bundled preparation derives depth from sparse structure-from-motion points
at roughly $0.01\%$ pixel coverage, which is not a credible dense benchmark.

\subsection{Backbones}
\label{sec:setup:backbones}
The primary frozen foundation is Depth Anything V2 Large \cite{Yang2024b}.
For the backbone ablation we additionally use MiDaS\,v3.x\,/\,DPT-Large
\cite{Ranftl2022,Ranftl2021}. Both are frozen. The competitor VI-Depth
\cite{Wofk2023} is run in its public configuration, \texttt{dpt-swin2-large-384}
with the SML head trained at $K{=}150$; its strongest published backbone,
\texttt{dpt-beit-large-512}, was unreachable at download time. Our method
and VI-Depth therefore differ in backbone, and VI-Depth's SML head is
evaluated below its training budget at $K{=}50$. We address both points
directly: a strict same-backbone, same-architecture control
(Sec.~\ref{sec:setup:baselines}) isolates the method from the backbone, and
a supplementary $K{=}150$ sweep at VI-Depth's own budget is reported in
Sec.~\ref{sec:discussion}.

\subsection{Baselines and the Five-Way Comparison}
\label{sec:setup:baselines}
Our headline protocol is a controlled five-way comparison on a single
evaluation harness: (i)~the published residual-on-CFA recipe on DAv2,
denoted \emph{vanilla}; (ii)~\textbf{B$'$}, the identical head trained with
random anchor-dropout augmentation---the sole configuration change is
\texttt{dropout\_rate\_max}${=}0.4$---which isolates implicit
augmentation-based robustness on the same backbone and architecture;
(iii)~MRAC applied to the vanilla head; (iv)~MRAC applied to the B$'$ head;
and (v)~VI-Depth official, plugged into the identical outlier-injection
harness. We further report a fixed-RANSAC \cite{Fischler1981} robust
residual---the obvious patch---and three closed-form CFA references
(ordinary least squares, Huber, and RANSAC) that anchor the absolute error
levels; deeper inlier-selection ablations appear in
Secs.~\ref{sec:substrate}--\ref{sec:diagnostics}. Published full-supervision
and depth-completion baselines (BTS \cite{Lee2019}, AdaBins \cite{Bhat2021},
ZoeDepth \cite{Bhat2023}, NLSPN \cite{Park2020}) are cited for context only
in Sec.~\ref{sec:robustness}.

\subsection{Metrics}
\label{sec:setup:metrics}
We report the standard metric-depth measures: absolute relative error
(AbsRel), squared relative error (SqRel), RMSE, RMSE-log, $\log_{10}$ error,
and threshold accuracies $\delta_i$ at $1.25^{\,i}$ for $i \in \{1,2,3\}$.
The maximum evaluation depth is $80$\,m. Headline comparisons report AbsRel
and $\delta_1$; complete metric tables are provided in the supplementary
material.

\subsection{Anchor Protocol}
\label{sec:setup:anchors}
The default budget is $K{=}50$ anchors per image, sampled uniformly at
random from valid-ground-truth pixels; we sweep $K \in [5,200]$ in
Sec.~\ref{sec:substrate}. The anchor draw is fixed by a per-image random
seed and is \emph{identical across all five methods}, so every method is
calibrated from the same measurements. The default protocol uses seed~$0$,
and the eight headline cells additionally report mean\,$\pm$\,std over seeds
$\{0,1,2\}$ (Sec.~\ref{sec:robustness}). Outliers are injected by replacing
a fraction $p \in \{0, 0.1, 0.25, 0.4\}$ of the anchor depths with one of
the four sensor-grounded processes of Sec.~\ref{sec:method:outliers}. The
full evaluation is therefore a grid of $4$ datasets $\times$ $4$ outlier
types $\times$ $4$ fractions $\times$ $5$ methods $=320$ cells.

% =====================================================================
\section{Anchor-Outlier Robustness}
\label{sec:robustness}

This section reports the headline result. We first show that the published
residual-on-CFA recipe collapses under sensor outliers and that
augmentation-based robustness does not rescue it
(Sec.~\ref{sec:rob:collapse}); we then show that the strongest deployed
competitor, VI-Depth, is robust to \emph{missing} anchors but has a blind
spot on \emph{present-but-wrong} multipath anchors
(Sec.~\ref{sec:rob:blindspot}); and we show that MRAC, a parameter-free
inference-time wrapper, repairs both failure modes
(Secs.~\ref{sec:rob:dominate}--\ref{sec:rob:crossbackbone}). All AbsRel
values are means over three seeds; standard deviations are reported in
Tables~\ref{tab:dropout25} and~\ref{tab:near25}.

\subsection{The Recipe Collapses; Augmentation Barely Helps}
\label{sec:rob:collapse}
Table~\ref{tab:dropout25} reports AbsRel under $25\%$ dropout outliers, the
max-range/no-return failure of real range sensors. The faithful
residual-on-CFA re-implementation (vanilla) collapses: AbsRel reaches
$1.490$ on KITTI and $1.471$ on DIODE, an order of magnitude above its
clean accuracy. The same-backbone, same-architecture augmentation control
B$'$, whose only difference from vanilla is training-time random anchor
dropout, recovers almost nothing---$0.347\!\to\!0.336$ on NYUv2 ($\approx
3\%$) and $1.490\!\to\!1.479$ on KITTI ($<1\%$)---and the absolute error on
KITTI and DIODE remains catastrophic. Implicit augmentation does not scale
down to a compact $0.47$\,M-parameter residual head: the head lacks the
capacity to internalize an outlier-rejection mechanism from data alone.

\begin{table}[t]
\centering
\caption{AbsRel under $25\%$ dropout outliers (max-range / no-return).
Mean\,$\pm$\,std over $3$ seeds. Lower is better; best per column in bold.
Vanilla and B$'$ collapse; MRAC restores robust accuracy on the same
backbone with no added parameters.}
\label{tab:dropout25}
\footnotesize
\setlength{\tabcolsep}{1pt}
\begin{tabular}{lcccc}
\toprule
Method & NYUv2 & KITTI & DIODE & SUN RGB-D \\
\midrule
vanilla (DAv2) & 0.347{\tiny$\pm$0.001} & 1.490{\tiny$\pm$0.005} & 1.471{\tiny$\pm$0.003} & 0.429{\tiny$\pm$0.001} \\
B$'$ (DAv2) & 0.336{\tiny$\pm$0.002} & 1.479{\tiny$\pm$0.005} & 1.432{\tiny$\pm$0.003} & 0.411{\tiny$\pm$0.001} \\
MRAC/van (ours) & 0.105{\tiny$\pm$0.001} & \textbf{0.125}{\tiny$\pm$0.001} & 0.348{\tiny$\pm$0.009} & 0.161{\tiny$\pm$0.001} \\
MRAC/B$'$ (ours) & \textbf{0.100}{\tiny$\pm$0.001} & 0.128{\tiny$\pm$0.001} & 0.340{\tiny$\pm$0.008} & \textbf{0.151}{\tiny$\pm$0.001} \\
VI-Depth & 0.144{\tiny$\pm$0.001} & 0.174{\tiny$\pm$0.001} & \textbf{0.323}{\tiny$\pm$0.002} & 0.195{\tiny$\pm$0.000} \\
\bottomrule
\end{tabular}
\end{table}

The obvious robust patch fails for the same reason Sec.~\ref{sec:method:outliers}
anticipates. Replacing the residual head's internal fit with a
fixed-threshold RANSAC \cite{Fischler1981} robust residual makes dropout
\emph{worse}, not better: on KITTI it reaches $2.894 \pm 0.058$ at $25\%$
dropout and $6.082 \pm 0.008$ at $40\%$---above even the unprotected vanilla
head---because the correlated dropout anchors form a self-consistent
pseudo-inlier set that the consensus step selects. The same RANSAC patch is
unremarkable on the uncorrelated families (KITTI uniform $0.198$, mixed
$0.220$ at $40\%$), confirming that correlation, not outlier magnitude, is
what defeats it.

\subsection{The Strongest Competitor Has a Multipath Blind Spot}
\label{sec:rob:blindspot}
VI-Depth tells the opposite half of the story. On dropout
(Table~\ref{tab:dropout25}) its per-$K$ SML head is genuinely robust
($0.144$--$0.323$), because anchor-dropout is precisely the corruption its
training simulates. But on $25\%$ near (multipath/specular) outliers
(Table~\ref{tab:near25}) it fails: AbsRel rises to $0.250$ on NYUv2 and
$0.489$ on KITTI. VI-Depth is beaten by the \emph{unprotected} vanilla head
on three of the four datasets and ties it on SUN RGB-D
($0.244$ vs $0.244$); on KITTI it is $2.9\times$ worse than vanilla and
$3.2\times$ worse than MRAC. The mechanism is structural: implicit
anchor-dropout training teaches a model to tolerate anchors that are
\emph{missing}, but a multipath anchor is \emph{present with a wrong value},
and VI-Depth's pipeline contains no explicit step that can reject it. The
headline cell is KITTI/near/$25\%$: VI-Depth $0.489 \pm 0.002$, vanilla
$0.170 \pm 0.000$, B$'$ $0.164 \pm 0.001$, MRAC $0.151 \pm 0.002$---a
$3.2\times$ error reduction over the strongest deployed competitor, with
standard deviations two orders of magnitude below the gap.

\begin{table}[t]
\centering
\caption{AbsRel under $25\%$ near (multipath / specular) outliers---the
present-but-wrong-value regime. Mean\,$\pm$\,std over $3$ seeds. MRAC wins
every dataset; VI-Depth is the worst method on NYUv2 and KITTI.}
\label{tab:near25}
\footnotesize
\setlength{\tabcolsep}{1pt}
\begin{tabular}{lcccc}
\toprule
Method & NYUv2 & KITTI & DIODE & SUN RGB-D \\
\midrule
vanilla (DAv2) & 0.192{\tiny$\pm$0.001} & 0.170{\tiny$\pm$0.000} & 0.366{\tiny$\pm$0.001} & 0.244{\tiny$\pm$0.000} \\
B$'$ (DAv2) & 0.199{\tiny$\pm$0.001} & 0.164{\tiny$\pm$0.001} & 0.364{\tiny$\pm$0.001} & 0.249{\tiny$\pm$0.001} \\
MRAC/van (ours) & 0.109{\tiny$\pm$0.001} & \textbf{0.149}{\tiny$\pm$0.002} & 0.273{\tiny$\pm$0.004} & 0.160{\tiny$\pm$0.002} \\
MRAC/B$'$ (ours) & \textbf{0.105}{\tiny$\pm$0.001} & 0.151{\tiny$\pm$0.002} & \textbf{0.267}{\tiny$\pm$0.005} & \textbf{0.152}{\tiny$\pm$0.001} \\
VI-Depth & 0.250{\tiny$\pm$0.001} & 0.489{\tiny$\pm$0.002} & 0.421{\tiny$\pm$0.001} & 0.244{\tiny$\pm$0.001} \\
\bottomrule
\end{tabular}
\end{table}

\subsection{MRAC Dominates the Same-Backbone Comparison}
\label{sec:rob:dominate}
\begin{figure*}[t]
\centering
\includegraphics[width=\textwidth]{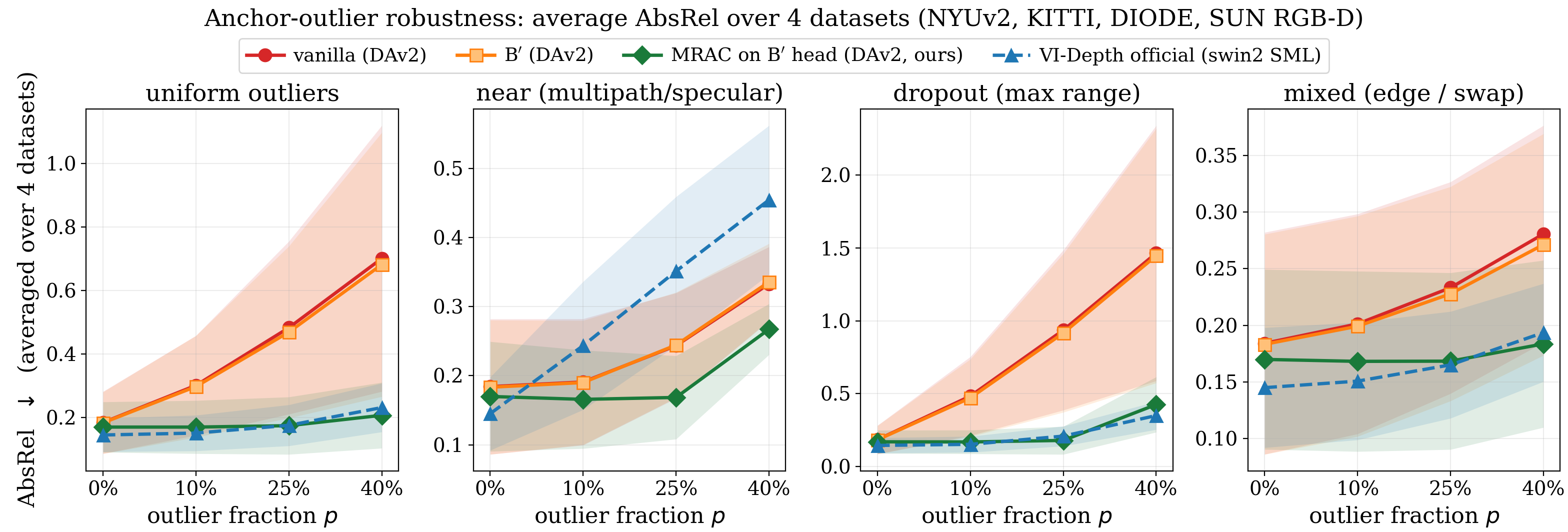}
\caption{AbsRel versus outlier fraction $p$ for the four outlier families,
averaged over NYUv2, KITTI, DIODE, and SUN RGB-D. Four methods: vanilla
(DAv2), B$'$ (DAv2), MRAC on the B$'$ head (ours), and VI-Depth official.
MRAC remains flat where the baselines diverge on dropout and where
VI-Depth's curve climbs on near (multipath). Shaded bands show $\pm 1$
standard deviation across the four datasets.}
\label{fig:robustness}
\end{figure*}

Tables~\ref{tab:dropout25} and~\ref{tab:near25} already show the per-cell
outcome; Table~\ref{tab:scorecard} aggregates it. Against its own substrate,
MRAC strictly wins $54$ of $64$ cells ($84\%$) whether the underlying head
is the vanilla head or the augmentation-trained B$'$ head, and the win
holds across all four outlier families. Two controls bound the result from
the other side: augmentation alone (B$'$ vs vanilla) wins only $52/64$ and
concentrates on dropout ($15/16$ dropout, $10/16$ near), and VI-Depth's
larger per-$K$ head beats B$'$ on $51/64$. MRAC closes the gap that
augmentation cannot, at zero added parameters. The gain is
\emph{mechanism-orthogonal}: applying MRAC on top of the
augmentation-trained head reaches the best or near-best AbsRel in both tables, so the
robust selection and the implicit augmentation are complementary rather
than redundant. Fig.~\ref{fig:robustness} plots AbsRel against outlier
fraction for the four methods across all four families, averaged over the
four datasets; the MRAC curves remain flat where vanilla and B$'$ diverge
and where VI-Depth's near curve climbs steeply.

\begin{table}[t]
\centering
\caption{Strict per-cell win counts on the seed-$0$ grid ($64$ cells per
pairwise comparison: $4$ datasets $\times$ $4$ outlier types $\times$ $4$
fractions). MRAC's same-backbone gain ($84\%$) is invariant to which head
it wraps.}
\label{tab:scorecard}
\footnotesize
\setlength{\tabcolsep}{1pt}
\begin{tabular}{lcl}
\toprule
Comparison & Wins & By outlier type \\
\midrule
MRAC vs its vanilla head & 54/64 (84\%) & universal across types \\
MRAC vs its B$'$ head & 54/64 (84\%) & gain orthogonal to aug. \\
B$'$ vs vanilla & 52/64 (81\%) & drop $15/16$; near $10/16$ \\
VI-Depth vs B$'$ & 51/64 (80\%) & scale $+$ per-$K$ training \\
MRAC vs VI-Depth & 34/64 (53\%) & near $13/16$; KITTI $16/16$ \\
\bottomrule
\end{tabular}
\end{table}

\subsection{Mechanism Orthogonality and \texorpdfstring{$K$}{K}-Agnosticism}
\label{sec:rob:kagnostic}
MRAC is an inference-time wrapper with no learned parameters, so it inherits
two deployment properties that the per-$K$ SML pipeline cannot offer. First,
because it adds no training, it improves any already-trained residual-on-CFA
head by the same selection mechanism---the $84\%$ win in
Table~\ref{tab:scorecard} is identical for the vanilla and B$'$ substrates.
Second, MRAC is $K$-agnostic: the Theil--Sen fit and MAD gate are defined for
any $K \ge 2$, and the single shared head serves $K \in [5,200]$
(Sec.~\ref{sec:substrate}). VI-Depth's public release instead ships three
separate SML checkpoints at $K = 150, 500, 1500$; changing the anchor budget
requires changing checkpoints, and any budget outside those three values is
out of distribution for its heads. For a sensor that delivers a handful of
anchors, MRAC's accuracy edge on multipath and its freedom from per-$K$
retraining are two facets of the same parameter-free design.

\subsection{Cross-Backbone Comparison Against VI-Depth}
\label{sec:rob:crossbackbone}
Against VI-Depth, MRAC runs on a different backbone (DAv2 vs
\texttt{dpt-swin2-large-384}), so the comparison is not backbone-matched;
the same-backbone result above is B$'$. With that caveat, MRAC wins $34$ of
$64$ cells overall ($53\%$), and the losses are concentrated exactly where
VI-Depth's stronger backbone is expected to win: the clean and low-$p$ cells
on DIODE and SUN RGB-D. Where the corruption is the failure mode that
matters, MRAC dominates. It wins $13/16$ near cells ($81\%$); the only three
near cells it loses are the clean $p{=}0$ cells on NYUv2, DIODE, and
SUN RGB-D, so MRAC wins \emph{all twelve cells in which multipath corruption
is actually present}. On KITTI, where DAv2 is in domain, MRAC wins all $16$
cells on the seed-$0$ grid, including the $3.2\times$ multipath gap of
Sec.~\ref{sec:rob:blindspot}. We attribute VI-Depth's DIODE/SUN RGB-D clean
wins to its backbone, not its calibration, and discuss the backbone
confound in Sec.~\ref{sec:discussion}.

\subsection{Operating Limits}
\label{sec:rob:limits}
At $40\%$ outliers the inlier fraction falls below the Theil--Sen breakdown
point of $\approx 29\%$, and MRAC degrades; we report this honestly. On
KITTI dropout at $40\%$, vanilla and B$'$ are unusable ($2.420 \pm 0.007$
and $2.416 \pm 0.007$) and the RANSAC patch is far worse ($6.082 \pm
0.008$), while MRAC stays at $0.324 \pm 0.036$ and VI-Depth at
$0.314 \pm 0.006$. MRAC and VI-Depth are therefore a statistical tie at this
cell: MRAC wins it on two of three seeds ($0.314, 0.294$) but loses the
third ($0.363$), and the resulting $\pm0.036$ standard deviation---against
$\approx \pm0.001$ everywhere else---is the breakdown signature, the trace
of draws that leave too few true inliers for the Theil--Sen median to
recover. This is the single headline cell whose outcome lies within seed
noise; it is why we state $16/16$ on the seed-$0$ grid rather than as a
seed-robust claim. The breakdown is dropout-specific: on KITTI near at
$40\%$ the order does not invert against VI-Depth ($0.299 \pm 0.004$ vs
$0.596 \pm 0.001$, a $2\times$ MRAC win), though here the less-destructive
near outliers leave the unprotected vanilla head ($0.275$) marginally ahead
of MRAC, the same effect that produces MRAC's two scorecard losses at KITTI
near. Within the $\le 29\%$ regime that the Theil--Sen guarantee covers, and
which spans the realistic sensor-outlier rates this paper targets, MRAC is
robust; past it, every method degrades and MRAC remains competitive with the
best.

% =====================================================================
\section{Calibration Substrate Study}
\label{sec:substrate}

MRAC operates on top of a calibration substrate it does not modify. This
section characterizes that substrate---its zero-shot behavior, its operating
range in the anchor budget $K$, its insensitivity to anchor placement, and
its response to non-outlier noise---to establish why residual-on-CFA is the
right object to make robust.

\subsection{Zero-Shot Generalization}
\label{sec:sub:zeroshot}
Table~\ref{tab:zeroshot} reports clean ($p{=}0$) AbsRel for a raw-depth
learned calibrator (a compact U-Net regressing depth directly from the image,
relative depth, and anchors, with no closed-form substrate), the three
closed-form CFA variants, the residual-on-CFA
architecture, and MRAC. The first observation is that a learned calibrator
\emph{without} the closed-form substrate is worse than a closed-form affine
fit in the zero-shot regime: the raw-depth head reaches $0.484$ on DIODE and
$0.301$ on SUN RGB-D, behind cfa-ransac ($0.321$) and cfa-huber ($0.166$)
respectively. A small network given the anchors directly does not, on its
own, recover a reliable metric scale across domains; the closed-form fit is
the more robust starting point.

The residual-on-CFA architecture closes most of this gap. By predicting a
correction to the closed-form fit rather than the depth itself, it improves
on every CFA variant on NYUv2 ($0.104$), KITTI ($0.133$), and SUN RGB-D
($0.149$); on DIODE the residual alone ($0.351$) still trails cfa-ransac
($0.321$), leaving one dataset where the substrate is not yet sufficient.

MRAC adds negligible clean-accuracy cost, and on the augmentation-trained
head it removes the remaining DIODE deficit. On the B$'$ head, MRAC improves
KITTI ($0.146\!\to\!0.128$) and DIODE ($0.348\!\to\!0.304$)---the only DAv2
configuration to beat cfa-ransac on DIODE---and stays within
$0.003$--$0.005$ AbsRel of the best DAv2 number on NYUv2 ($0.100$ vs
$0.097$) and SUN RGB-D ($0.147$ vs $0.142$). On the vanilla head, MRAC
slightly regresses clean NYUv2 ($0.104\!\to\!0.111$) and SUN RGB-D
($0.149\!\to\!0.163$) while still improving KITTI and DIODE. Robustness is
therefore not purchased at the cost of clean accuracy: the robust selection
that repairs the outlier behavior of Sec.~\ref{sec:robustness} is at worst
mildly conservative on clean data and is net beneficial on the
augmentation-trained substrate. VI-Depth attains the lowest clean AbsRel on
NYUv2 ($0.079$), DIODE ($0.218$), and SUN RGB-D ($0.114$); we attribute this
to its swin2 backbone rather than its calibration, consistent with
Sec.~\ref{sec:rob:crossbackbone} and discussed in Sec.~\ref{sec:discussion}.

\begin{table}[t]
\centering
\caption{Clean ($p{=}0$) AbsRel. The raw-depth learned calibrator loses to a
closed-form fit on DIODE and SUN RGB-D; residual-on-CFA rescues most cells;
MRAC adds negligible clean cost and removes the DIODE deficit on the B$'$
head. Best per column in bold. VI-Depth uses a different (swin2) backbone.}
\label{tab:zeroshot}
\footnotesize
\setlength{\tabcolsep}{2pt}
\begin{tabular}{lcccc}
\toprule
Method & NYUv2 & KITTI & DIODE & SUN RGB-D \\
\midrule
raw-depth head (no CFA) & 0.207 & 0.175 & 0.484 & 0.301 \\
\midrule
cfa-ols & 0.155 & 0.333 & 0.386 & 0.173 \\
cfa-huber & 0.144 & 0.233 & 0.339 & 0.166 \\
cfa-ransac & 0.155 & 0.204 & 0.321 & 0.167 \\
\midrule
residual (vanilla) & 0.104 & 0.133 & 0.351 & 0.149 \\
B$'$ (dropout-aug) & 0.097 & 0.146 & 0.348 & 0.142 \\
MRAC/van (ours) & 0.111 & \textbf{0.126} & 0.314 & 0.163 \\
MRAC/B$'$ (ours) & 0.100 & 0.128 & 0.304 & 0.147 \\
\midrule
VI-Depth (swin2) & \textbf{0.079} & 0.169 & \textbf{0.218} & \textbf{0.114} \\
\bottomrule
\end{tabular}
\end{table}

\subsection{Anchor-Budget Operating Curve}
\label{sec:sub:kcurve}
\begin{figure}[t]
\centering
\includegraphics[width=\columnwidth]{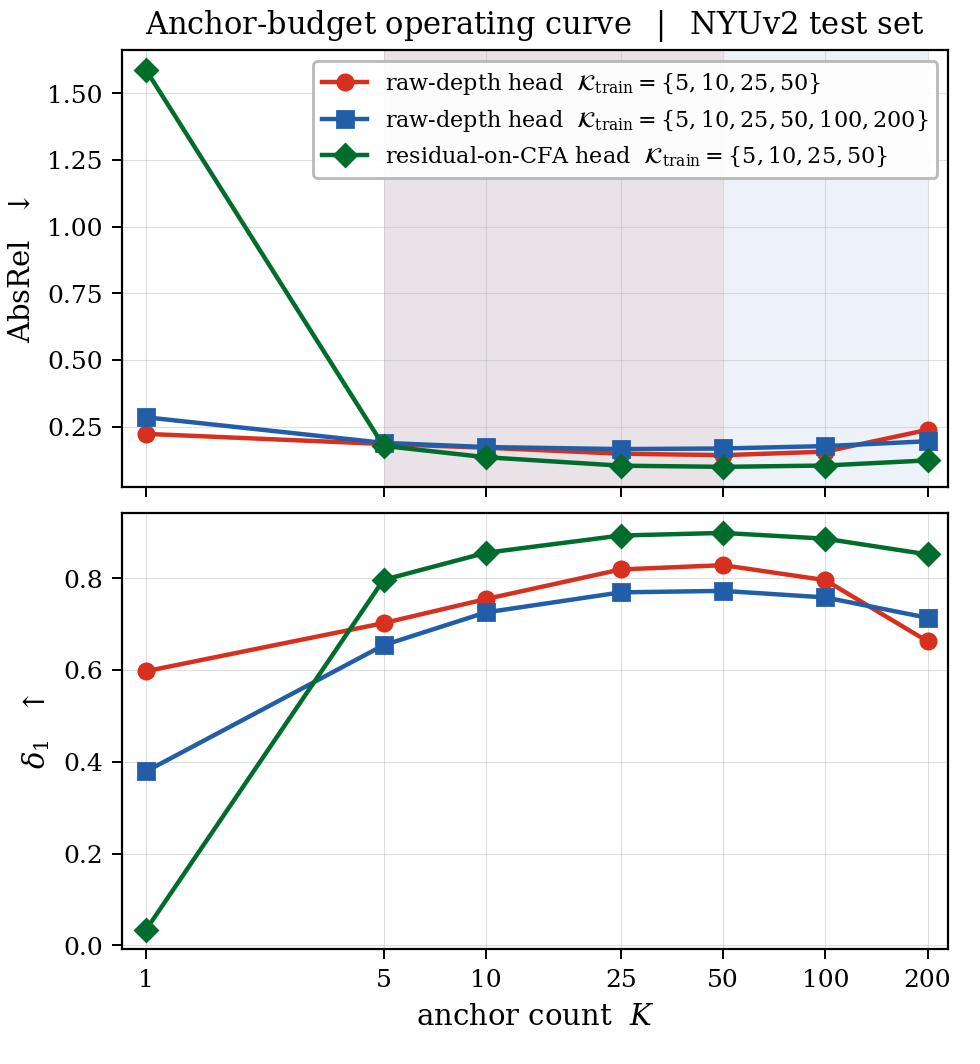}
\caption{AbsRel versus anchor budget $K$ on NYUv2 for three architectures.
Residual-on-CFA dominates for $K \ge 5$; $K{=}1$ is the architectural
degenerate point (the closed-form fit needs $K \ge 2$); the curve is
U-shaped with a sweet spot at $K^{\!*}{=}50$.}
\label{fig:kcurve}
\end{figure}

Fig.~\ref{fig:kcurve} overlays the AbsRel-vs-$K$ curves of three
architectures on NYUv2; the residual-on-CFA curve dominates the alternatives
for all $K \ge 5$. The curve is U-shaped. At $K{=}1$ the architecture is
degenerate---the closed-form fit~\eqref{eq:cfa} requires $K \ge 2$ to
determine a slope---and AbsRel is $1.586$. Accuracy then improves sharply
($0.178$ at $K{=}5$, $0.135$ at $K{=}10$) to a sweet spot of $0.099$ at
$K^{\!*}{=}50$,\footnote{The clean NYUv2 AbsRel reported here ($0.099$ at
$K{=}50$) is from the $K$-sweep evaluation, whereas
Table~\ref{tab:zeroshot} reports $0.104$ for the same configuration from the
outlier-injection harness. These are independent evaluation runs with
independent per-image anchor draws; they differ by $\le 0.005$ AbsRel at
$K{=}50$, within run-to-run anchor-sampling variance.} after which it rises
again ($0.104$ at $K{=}100$, $0.123$ at $K{=}200$) as the network begins to
overfit the denser anchor mask. The practical consequence is that a single
trained head is accurate across more than an order of magnitude in anchor
budget. This is the basis of the $K$-agnosticism claim of
Sec.~\ref{sec:rob:kagnostic}: MRAC introduces no $K$-dependent parameters,
and the substrate it wraps spans $K \in [5,200]$ from one checkpoint, in
contrast to the per-$K$ SML checkpoints of the deployed competitor.

\subsection{Anchor Placement and Pattern}
\label{sec:sub:pattern}
A natural question is whether the anchor \emph{placement} matters as much as
the anchor \emph{count}. Under the residual-on-CFA substrate it largely does
not: the substrate extracts depth structure from the global relative-depth
field, so it is insensitive to where the sparse anchors fall, and random
sampling is near-optimal. We present the controlled placement
evidence---random, farthest-point, uncertainty-active, and stratified sampling---in
Sec.~\ref{sec:diagnostics}, where uncertainty-active placement is shown to
\emph{hurt} monotonically. The implication for the present section is that
the substrate absorbs the placement degree of freedom, which is what makes
\emph{anchor trustworthiness}, rather than anchor location, the lever that
Sec.~\ref{sec:robustness} targets.

\subsection{Robustness to Gaussian Anchor Noise}
\label{sec:sub:noise}
Outlier robustness must be distinguished from robustness to ordinary
measurement noise. We perturb every anchor depth with zero-mean Gaussian
noise of standard deviation $\sigma$ and sweep $\sigma \in
[0.05, 0.50]$\,m. AbsRel is essentially flat across this range---moving from
the clean value only at the smallest $\sigma$ and then remaining constant to
within $0.001$ AbsRel out to $\sigma{=}0.5$\,m---because averaging $K{=}50$
zero-mean perturbations into the affine fit cancels the noise by the central
limit theorem. Outliers are categorically different: they are not zero-mean,
do not average out, and a single one can dominate a least-squares fit
(Sec.~\ref{sec:method:cfa}). This is precisely why Gaussian-noise robustness
is cheap and outlier robustness is not, and why this paper studies the
latter as a separate problem requiring an explicit mechanism rather than
relying on anchor-count pooling.

% =====================================================================
\section{Diagnostic Findings}
\label{sec:diagnostics}

This section reports two diagnostics that inform deployment. Both were run
during the raw-depth training phase rather than the filled pipeline of
Secs.~\ref{sec:robustness}--\ref{sec:substrate}; we therefore report them as
\emph{ordinal and architectural} findings---rankings and sign effects that
the filled pipeline inherits---rather than as absolute numbers comparable to
the preceding tables.

\subsection{Anchor Placement Is a Red Herring}
\label{sec:diag:placement}
Table~\ref{tab:placement} sweeps four anchor-placement strategies on NYUv2:
uniform random sampling, farthest-point sampling, uncertainty-active
sampling, and stratified sampling. The result is counterintuitive. Random
sampling is at or near the best AbsRel at every budget, and stratified
sampling matches it to within $0.003$; the two ``clever'' geometric and
information-theoretic strategies do not help. Farthest-point sampling is
uniformly worse than random, and uncertainty-active sampling---placing
anchors where a model is least confident---is not merely worse but degrades
\emph{monotonically} as the budget grows, from $0.266$ at $K{=}5$ to $0.546$
at $K{=}100$, the opposite of the expected trend.

The explanation is the frozen foundation. Because $d_{\mathrm{rel}}$ already
encodes global scene structure, the marginal value of a geometrically
optimized anchor is small, and concentrating anchors in high-uncertainty
regions starves the affine fit of the well-spread support it needs,
producing a fit that is accurate locally and biased globally. The practical
message is that anchor \emph{placement} is not the lever: a deployed system
gains nothing from sophisticated anchor selection and should instead spend
its effort on anchor \emph{trustworthiness}, which is the failure mode
Sec.~\ref{sec:robustness} addresses. This subsumes the anchor-pattern
question of Sec.~\ref{sec:sub:pattern}: once the substrate is fixed, where
the anchors fall is second-order.

\begin{table}[t]
\centering
\caption{AbsRel by anchor-placement strategy on NYUv2 (raw-depth phase;
ordinal result). Random and stratified sampling are near-optimal;
farthest-point is worse; uncertainty-active sampling degrades monotonically
as $K$ grows. Best per column in bold.}
\label{tab:placement}
\footnotesize
\setlength{\tabcolsep}{5pt}
\begin{tabular}{lccccc}
\toprule
Strategy & $K{=}5$ & $K{=}10$ & $K{=}25$ & $K{=}50$ & $K{=}100$ \\
\midrule
random & 0.142 & \textbf{0.121} & 0.117 & \textbf{0.109} & 0.113 \\
farthest-point & 0.252 & 0.232 & 0.194 & 0.177 & 0.160 \\
uncertainty-active & 0.266 & 0.361 & 0.444 & 0.480 & 0.546 \\
stratified & \textbf{0.132} & 0.125 & \textbf{0.112} & 0.112 & \textbf{0.111} \\
\bottomrule
\end{tabular}
\end{table}

\subsection{Loss--Architecture Interaction}
\label{sec:diag:loss}
The second diagnostic concerns the training losses, and its conclusion is
that loss design is architecture-dependent: a term that helps one
parameterization can harm another. We ablate the loss terms of
Sec.~\ref{sec:method:training} on the raw-depth head (four terms) and on the
residual-on-CFA architecture (which adds the auxiliary \emph{res} term),
reported in Table~\ref{tab:lossablation}.

Three findings stand out. First, removing the scale-invariant log term
(SILog) is catastrophic on the raw-depth head, where NYUv2 AbsRel rises from
$0.143$ to $0.697$, and clearly harmful on the residual architecture
($0.096\!\to\!0.123$). Second, and most diagnostic, the confidence-weighting
term (CWA) \emph{reverses sign} across the two architectures. On the
raw-depth head, removing CWA \emph{improves} zero-shot accuracy
($0.143 \to 0.113$ on NYUv2); on the residual-on-CFA architecture, removing
the same term is catastrophic for zero-shot generalization, with cross-domain
AbsRel exploding on DIODE ($0.671$) and SUN RGB-D ($0.675$). The residual
parameterization changes what uncertainty weighting does: where the
raw-depth model is better off without it, the residual substrate depends on
it to remain calibrated off the training domain. Third, the structural-scale
(SSC) and feature-matching (FMP) terms are largely redundant under the
residual architecture---both ablations sit within $\approx 0.01$ AbsRel of
the full model---so they contribute little once the closed-form substrate is
in place. Finally, the residual model's auxiliary \emph{res} term is itself
harmful: removing it \emph{improves} accuracy ($0.096\!\to\!0.084$ on NYUv2,
$0.214\!\to\!0.139$ on KITTI), and this $-$res setting is the recommended
configuration carried into the filled pipeline of
Secs.~\ref{sec:robustness}--\ref{sec:substrate}.

The deployment implication is that loss recipes do not transfer across
calibration architectures, and that the residual-on-CFA substrate this paper
builds on specifically requires the confidence-weighting term that a
raw-depth model would discard. Anyone porting a loss configuration between
the two parameterizations should expect to re-tune CWA in particular.

\begin{table}[t]
\centering
\caption{Loss--architecture ablation (raw-depth phase; AbsRel). Top: the
original unconstrained-U-Net raw-depth head on NYUv2. Bottom:
residual-on-CFA on four datasets. Removing CWA improves the raw-depth model
but is catastrophic on the residual architecture (DIODE/SUN RGB-D), the
sign reversal of Sec.~\ref{sec:diag:loss}. The recommended configuration
drops the auxiliary \emph{res} term ($-$res), the best residual row.}
\label{tab:lossablation}
\footnotesize
\setlength{\tabcolsep}{4pt}
\begin{tabular}{lcccc}
\toprule
Config & NYUv2 & KITTI & DIODE & SUN RGB-D \\
\midrule
\multicolumn{5}{l}{\textit{Raw-depth head (original U-Net)}}\\
full & 0.143 & --- & --- & --- \\
$-$silog & 0.697 & --- & --- & --- \\
$-$cwa & 0.113 & --- & --- & --- \\
$-$ssc & 0.258 & --- & --- & --- \\
$-$fmp & 0.162 & --- & --- & --- \\
\midrule
\multicolumn{5}{l}{\textit{Residual-on-CFA}}\\
full & 0.096 & 0.214 & 0.336 & 0.160 \\
$-$silog & 0.123 & 0.324 & 0.368 & 0.170 \\
$-$cwa & 0.109 & 0.227 & 0.671 & 0.675 \\
$-$ssc & 0.100 & 0.239 & 0.353 & 0.162 \\
$-$fmp & 0.096 & 0.217 & 0.330 & 0.161 \\
$-$res (recommended) & \textbf{0.084} & \textbf{0.139} & \textbf{0.326} & \textbf{0.155} \\
\bottomrule
\end{tabular}
\end{table}

% =====================================================================
\section{Inference Cost and Deployment}
\label{sec:cost}

MRAC is designed for deployment, and its cost claim is concrete: it adds no
parameters and no second forward pass. Table~\ref{tab:latency} reports
per-image latency on a single NVIDIA A100, measured over $100$ iterations
after $10$ warmup iterations at $K{=}50$.

\begin{table}[t]
\centering
\caption{Per-image inference latency on one NVIDIA A100 (80\,GB), mean over
$100$ iterations ($10$ warmup), $K{=}50$. The frozen foundation dominates
end-to-end time. MRAC's anchor selection (Theil--Sen $+$ MAD) runs on CPU at
$\approx 50\,\mu$s, below the resolution of the end-to-end timing, and adds
no forward pass; the residual head is called once, on the cleaned anchors.}
\label{tab:latency}
\footnotesize
\setlength{\tabcolsep}{3pt}
\begin{tabular}{lcc}
\toprule
Component & NYUv2 ($480{\times}640$) & KITTI ($352{\times}1216$) \\
\midrule
Foundation (DAv2-L) & 24.2\,{\tiny$\pm$0.26} & 26.4\,{\tiny$\pm$0.09} \\
Residual head ($R_\theta$) & 10.5\,{\tiny$\pm$0.02} & 15.0\,{\tiny$\pm$0.01} \\
MRAC selection (CPU) & $\approx 0.05$ & $\approx 0.05$ \\
\midrule
End-to-end & 34.5\,{\tiny$\pm$0.03} & 41.0\,{\tiny$\pm$0.10} \\
Throughput (FPS) & 29.0 & 24.4 \\
\bottomrule
\end{tabular}
\end{table}

The frozen foundation is the bottleneck, at $24.2$\,ms on NYUv2 and
$26.4$\,ms on KITTI, with the $0.47$\,M-parameter residual head adding
$10.5$ and $15.0$\,ms respectively. End-to-end latency is $34.5$\,ms
($29.0$\,FPS) on NYUv2 and $41.0$\,ms ($24.4$\,FPS) on KITTI---real time on a
single accelerator. Because the foundation dominates, any faster foundation
translates directly into a faster pipeline.

MRAC's overhead over the vanilla pipeline is the Theil--Sen fit and the MAD
gate of Sec.~\ref{sec:method:mrac}. The Theil--Sen slope is $O(K^2)$ in the
anchor pairs---about $K(K{-}1)/2 \approx 1.2\text{k}$ slopes at $K{=}50$---and
the MAD gate is $O(K)$; together they cost $\approx 50\,\mu$s on CPU, four
orders of magnitude below the end-to-end time and below what
Table~\ref{tab:latency} can resolve. The decisive point is architectural:
MRAC \emph{replaces} the anchor set passed to the residual head and calls
that head exactly once, so it introduces no second forward pass. Its
end-to-end latency therefore equals the vanilla pipeline's to within
measurement noise---robustness is free at inference. A naive alternative
that ran the head twice, once to detect and once to recalibrate, would add a
full head forward ($10$--$15$\,ms, a $30$--$40\%$ latency increase); MRAC
avoids this by performing all outlier reasoning in the $50\,\mu$s CPU step.

Two further properties matter for deployment. First, MRAC adds zero learned
parameters; the only trainable component is the $0.47$\,M-parameter head,
which is trained once and shared. Second, MRAC is $K$-agnostic: a single
checkpoint serves $K \in [5,200]$ (Sec.~\ref{sec:sub:kcurve}), so a deployed
system whose anchor budget varies with range, scene, or sensor mode needs no
per-budget retraining or checkpoint switching. The deployed competitor
instead ships three separate SML heads at $K = 150, 500, 1500$; matching an
arbitrary anchor budget requires selecting and loading the nearest head, and
budgets far from those values are out of distribution. For an embedded
metric-depth stack, MRAC is a single small head plus a $50\,\mu$s CPU
routine that delivers the outlier robustness of Sec.~\ref{sec:robustness}
without enlarging the model, the latency, or the checkpoint inventory.

% =====================================================================
\section{Discussion and Limitations}
\label{sec:discussion}

\subsection{Why the Blind Spot Exists}
\label{sec:disc:why}
VI-Depth's scale-map learner predicts a per-pixel scale and shift field
conditioned on the sparse anchors, and is trained with augmentation that
varies anchor count and placement---in effect simulating missing anchors.
This is exactly why it is robust to dropout (Sec.~\ref{sec:robustness}): a
missing anchor lies inside its training distribution. But its conditioning
contains no step that tests whether a \emph{present} anchor's value is
consistent with the scene geometry the foundation already encodes. A
multipath anchor is present and carries a plausible near-biased depth, so
the scale-map learner absorbs it and propagates the error spatially---an
effect measured in the supplement (Fig.~S1), where AbsRel decays
monotonically with distance from the corrupted anchor for VI-Depth and
stays flat for MRAC. The blind spot is therefore structural rather than a
tuning deficiency: the pipeline has no present-but-wrong-value rejection
mechanism.

\begin{figure*}[!t]
\centering
\includegraphics[width=\textwidth]{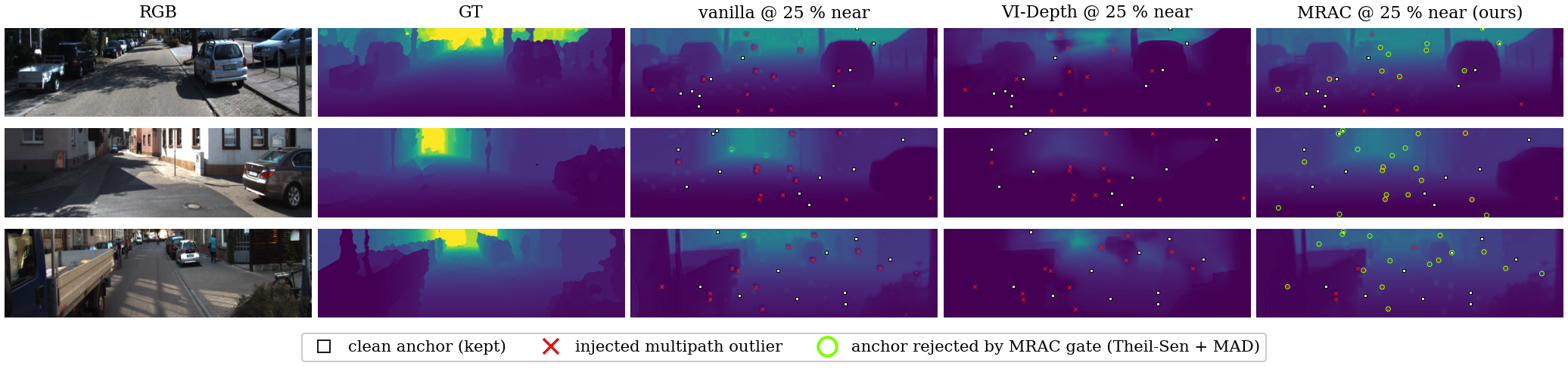}
\caption{Mechanism of the multipath blind spot on three KITTI frames under
$25\%$ near (multipath) outliers. Columns: RGB, ground truth, the vanilla
residual head, VI-Depth, and MRAC. Small white squares mark clean anchors
that are kept (subsampled to a representative $\sim 10$ per panel); red
crosses mark injected multipath outliers; green circles in the MRAC column
mark anchors rejected by the Theil--Sen $+$ MAD gate before the single
head call. VI-Depth absorbs the present-but-wrong-value anchors into its
scale field and distorts the surrounding depth; MRAC rejects most of them
and stays faithful to the scene. Red crosses without green circles in the
MRAC column are outliers the gate missed. Harness-wide gate P/R on this
cell (KITTI, near, $25\%$, $\kappa\!=\!2$): $0.60/0.64$.}
\label{fig:blindspot}
\end{figure*}

Foundation consistency is that missing mechanism. By testing each anchor
against the relative-depth ordering through the Theil--Sen fit
(Sec.~\ref{sec:method:mrac})---an independent witness the scale-map learner
never consults---MRAC rejects anchors that violate the foundation's geometry
regardless of whether the corruption is missing or present-with-wrong-value.
Fig.~\ref{fig:blindspot} makes this concrete: on multipath-corrupted KITTI
frames, VI-Depth's depth is distorted in the neighborhood of the corrupted
anchors, while MRAC---having gated those anchors out before the single head
call---remains faithful to the scene. The augmentation that makes VI-Depth
robust to missing anchors is silent on present-wrong ones; the
foundation-consistency gate that makes MRAC robust to present-wrong anchors
is, by Sec.~\ref{sec:robustness}, equally effective on missing ones.

\subsection{Limitations}
\label{sec:disc:limits}
We state the study's limitations plainly.

\textbf{Backbone.} Our method runs on DAv2-Large and VI-Depth on
\texttt{dpt-swin2-large-384}, so the cross-backbone comparison
(Sec.~\ref{sec:rob:crossbackbone}) confounds method and backbone. We close
this three ways. The load-bearing strength claim is the same-backbone,
same-architecture comparison against vanilla and B$'$ on DAv2, where MRAC's
only changed variable is inlier selection and it wins $84\%$ of cells
(Sec.~\ref{sec:rob:dominate})---establishing the method's value with no
backbone comparison at all. The multipath blind spot is structural and
reproduces at VI-Depth's own training budget (below), so it is not a backbone
artifact. And we attribute VI-Depth's clean-cell wins
(Sec.~\ref{sec:substrate}) to its backbone rather than its calibration, and
do not claim them. VI-Depth's strongest published backbone
(\texttt{dpt-beit-large-512}) was unreachable at download time; a stronger
backbone would likely improve its clean accuracy, but it cannot supply the
present-wrong rejection the pipeline structurally lacks.

\textbf{Anchor-budget mismatch.} VI-Depth's head is trained at $K{=}150$ and
we evaluate it at the $K{=}50$ operating point our $K$-agnosticism argument
targets. To rule out a mismatch artifact, we re-evaluate it at its own
$K{=}150$ on the multipath cells. The blind spot persists and, if anything,
widens: near/$25\%$ AbsRel moves from $0.250$ to $0.271$ on NYUv2, $0.420$ to
$0.432$ on DIODE, and $0.243$ to $0.258$ on SUN RGB-D, and is unchanged on
KITTI ($0.491$ to $0.490$). The multipath failure is therefore not a
consequence of evaluating below the training budget.

\textbf{Breakdown.} MRAC inherits the Theil--Sen breakdown point of
$\approx 29\%$; beyond $40\%$ outliers it degrades, and KITTI/dropout/$40\%$
is a statistical tie with VI-Depth (Sec.~\ref{sec:rob:limits}). This bounds
the guarantee to the realistic sensor-outlier rates this paper targets; past
it, no method we tested is reliable.

\textbf{Gate diagnostics.} Fig.~\ref{fig:blindspot} concedes that the MAD
gate misses some outliers; here we quantify those misses. On the eight
headline cells at
$\kappa\!=\!2$ the gate has mean precision $0.75$ and mean recall $0.83$
against the injected-outlier ground truth. Dropout is essentially perfect
(recall $\ge 0.90$ on three datasets and $1.00$ on KITTI); near is weaker,
and the hardest cell is KITTI near, where the driving-scene depth spread
gives precision $0.60$ and recall $0.64$: $40\%$ of clean anchors are
falsely rejected and $36\%$ of injected multipath outliers are missed,
which is precisely the
visible gap between MRAC and vanilla on that cell. The residual error
propagates spatially (supplement Fig.~S1). Per-cell precision and recall
are listed in the supplement.

\textbf{Protocol.} The strict-win scorecard is computed on a single
(seed-$0$) realization of the anchor draw, and per-cell counts can shift
slightly under reseeding, as the KITTI/dropout/$40\%$ tie illustrates; the
headline tables carry $3$-seed bars. SUN RGB-D is evaluated on a stride-$5$
subset, since the official $5050$-image split requires toolbox files absent
from the public release; ETH3D is excluded for inadequate dense coverage;
and NYUv2 uses the standard filled-depth protocol, applied identically to
all methods.

\subsection{Future Work}
\label{sec:disc:future}
Two directions follow. Foundation consistency is presently a hand-built
robust fit; a learned consistency model that scores anchors against the
relative-depth field could be folded into the head at the same
single-forward-pass cost, particularly for the multipath cells where the
MAD gate's recall trails its dropout ceiling. The wrapper is
foundation-agnostic by construction, so applying it to other frozen
geometry foundations---MoGe-2, Marigold, Depth Pro---is immediate and
would test whether the blind spot is specific to VI-Depth's scale-map
learner or general to anchor-conditioned calibration.

% =====================================================================
\section{Conclusion}
\label{sec:conclusion}

Sparse-anchor metric depth calibration from frozen foundations has
assumed clean anchors, yet real range sensors produce outliers that are
present with the wrong value, not merely missing. We showed that the
published residual-on-CFA recipe collapses under such outliers, and that
VI-Depth, the strongest publicly deployed sparse-anchor method, has a
structural multipath blind spot: its anchor-conditioned scale map has no
mechanism to reject present-but-wrong-value anchors.

MRAC, our proposed parameter-free inference-time wrapper, gates anchors
by foundation consistency (Theil--Sen + MAD) before a single call to the
residual-on-CFA head, adds no learned parameters, and runs at
$\approx 50\,\mu$s on CPU. It strictly wins $84\%$ of the same-backbone
cells and, against VI-Depth, all twelve corrupted multipath cells and all
sixteen KITTI cells, reducing KITTI multipath AbsRel by $3.2\times$
($0.489\!\to\!0.151$) with no retraining, and serves $K \in [5,200]$ from
one checkpoint. The broader lesson is that robustness to missing anchors
and robustness to wrong-valued anchors are distinct problems, and that
the foundation's relative-depth geometry is a cheap, sufficient witness
for the second.

% =====================================================================
% Scriptsize bibliography — TPAMI camera-ready pattern; keeps the paper
% at 12 pages with the added kappa / gate-diagnostics paragraphs.
% We patch \thebibliography so IEEEtran's internal font choice inherits
% the outer size, and drop \itemsep for a compact list.
\let\OldThebib\thebibliography
\renewcommand{\thebibliography}[1]{%
  \OldThebib{#1}%
  \scriptsize
  \setlength{\itemsep}{0pt plus 0.1ex}%
  \setlength{\parsep}{0pt}%
}
\section*{Acknowledgment}
The authors gratefully acknowledge the support from the Department of Science
and Technology (DST), Government of India, through the DST-FIST grant
[Sanction No.: SR/FST/MS-I/2022/116].

\bibliographystyle{IEEEtran}
\bibliography{references}

\clearpage
\onecolumn\twocolumn[\begin{center}\LARGE\bfseries Supplementary Material\end{center}\vspace{6pt}]
\setcounter{section}{0}\setcounter{table}{0}\setcounter{figure}{0}

% Supplement figures/tables numbered with an "S" prefix so main-paper
% cross-references ("supplement Fig.~S1", "supplement Table~SI") are
% unambiguous.
\renewcommand{\thefigure}{S\arabic{figure}}
\renewcommand{\thetable}{S\Roman{table}}

\section{Sensitivity to the MAD multiplier $\kappa$}
\label{sec:supp:kappa}

Section~III-E of the main paper fixes $\kappa\!=\!2$ and reports a
sensitivity sweep on the eight headline cells (four datasets, near and
dropout at $25\%$). Table~\ref{tab:kappa_absrel} gives the per-cell
AbsRel numbers behind that summary. All rows are MRAC on the B$'$
head. The paper's headline numbers (Table~I and Table~II) correspond
to the $\kappa\!=\!2.0$ column and agree with it to within
$\pm 0.008$ AbsRel across the eight cells, the run-to-run anchor-sampling
variance.

Two observations. First, $\kappa\!=\!1.5$ is marginally best on seven
of the eight cells: by at most $0.004$ AbsRel on six of them and by
$0.015$ on DIODE dropout. The mean cost of the paper's chosen
$\kappa\!=\!2$ across the eight cells is therefore about $0.004$ AbsRel,
with the compensating benefit of higher gate precision
(Table~\ref{tab:kappa_gate}). Second, KITTI dropout is essentially
invariant to $\kappa$ over the swept range because the injected
outliers (sensor max range, $\approx 80\,$m) sit so far from the
Theil--Sen residual median that the gate identifies them at every
$\kappa$; the observed movement is entirely at the second decimal.
The hardest cell is DIODE dropout, where the wide outdoor depth range
gives the MAD gate a broader residual distribution and $\kappa\!=\!1.5$
outperforms $\kappa\!=\!3.0$ by $0.039$ AbsRel.

\begin{table*}[t]
\centering
\caption{MRAC AbsRel across $\kappa$, per cell. Lower is better. Best
per row in bold. All rows are MRAC on the B$'$ head at $K\!=\!50$,
$25\%$ outlier fraction, seed 0.}
\label{tab:kappa_absrel}
\footnotesize
\setlength{\tabcolsep}{6pt}
\begin{tabular}{llcccc}
\toprule
Dataset  & Cell     & $\kappa\!=\!1.5$ & $\kappa\!=\!2.0$ & $\kappa\!=\!2.5$ & $\kappa\!=\!3.0$ \\
\midrule
NYUv2    & near     & \textbf{0.104} & 0.106 & 0.109 & 0.113 \\
NYUv2    & dropout  & \textbf{0.097} & 0.101 & 0.104 & 0.108 \\
KITTI    & near     & \textbf{0.148} & 0.152 & 0.157 & 0.165 \\
KITTI    & dropout  & 0.128          & \textbf{0.127} & 0.128 & 0.128 \\
DIODE    & near     & \textbf{0.267} & 0.269 & 0.271 & 0.273 \\
DIODE    & dropout  & \textbf{0.333} & 0.348 & 0.359 & 0.372 \\
SUN RGB-D & near    & \textbf{0.148} & 0.150 & 0.153 & 0.157 \\
SUN RGB-D & dropout & \textbf{0.147} & 0.150 & 0.154 & 0.158 \\
\bottomrule
\end{tabular}
\end{table*}

\section{MAD-gate precision and recall}
\label{sec:supp:gate}

Section~IX-B of the main paper reports mean gate precision $0.75$ and
mean recall $0.83$ at $\kappa\!=\!2$ over the eight headline cells and
identifies KITTI near as the hardest cell. Table~\ref{tab:kappa_gate}
gives the per-cell precision and recall for the full $\kappa$ sweep.

Precision (fraction of MAD-rejected anchors that were true injected
outliers) rises monotonically with $\kappa$; recall (fraction of
injected outliers actually rejected) falls monotonically. The two
cross near $\kappa\!\approx\!2$ for most cells, which is the empirical
justification for the paper's default. KITTI dropout is the exception:
recall stays at $1.00$ across the sweep because Theil--Sen's median
slope is essentially undisturbed by max-range corruption, so precision
alone is left to move. KITTI near is the hardest cell, matching the
qualitative gate misses visible as bare red crosses in Fig.~5 of the
main paper.

\begin{table*}[t]
\centering
\caption{MAD-gate precision / recall vs $\kappa$, per cell. Each entry
is $P / R$ where $P$ = fraction of MAD-rejected anchors that were true
injected outliers, $R$ = fraction of injected outliers actually
rejected. Best precision per row in bold; best recall per row underlined.}
\label{tab:kappa_gate}
\footnotesize
\setlength{\tabcolsep}{6pt}
\begin{tabular}{llcccc}
\toprule
Dataset  & Cell     & $\kappa\!=\!1.5$ & $\kappa\!=\!2.0$ & $\kappa\!=\!2.5$ & $\kappa\!=\!3.0$ \\
\midrule
NYUv2    & near     & 0.73 / \underline{0.86} & 0.77 / 0.82 & 0.80 / 0.77 & \textbf{0.83} / 0.72 \\
NYUv2    & dropout  & 0.80 / \underline{0.88} & 0.84 / 0.83 & 0.86 / 0.78 & \textbf{0.88} / 0.73 \\
KITTI    & near     & 0.60 / \underline{0.73} & 0.60 / 0.64 & \textbf{0.60} / 0.55 & 0.58 / 0.46 \\
KITTI    & dropout  & 0.74 / \underline{1.00} & 0.77 / \underline{1.00} & 0.79 / \underline{1.00} & \textbf{0.81} / \underline{1.00} \\
DIODE    & near     & 0.65 / \underline{0.75} & 0.68 / 0.70 & 0.70 / 0.66 & \textbf{0.72} / 0.62 \\
DIODE    & dropout  & 0.74 / \underline{0.96} & 0.78 / 0.94 & 0.81 / 0.93 & \textbf{0.83} / 0.91 \\
SUN RGB-D & near    & 0.72 / \underline{0.86} & 0.76 / 0.81 & 0.79 / 0.77 & \textbf{0.82} / 0.72 \\
SUN RGB-D & dropout & 0.78 / \underline{0.93} & 0.81 / 0.90 & 0.84 / 0.87 & \textbf{0.86} / 0.84 \\
\bottomrule
\end{tabular}
\end{table*}

\section{Mechanism figure: spatial propagation of the multipath error}
\label{sec:supp:mechanism}

\begin{figure*}[!t]
\centering
\includegraphics[width=0.72\textwidth]{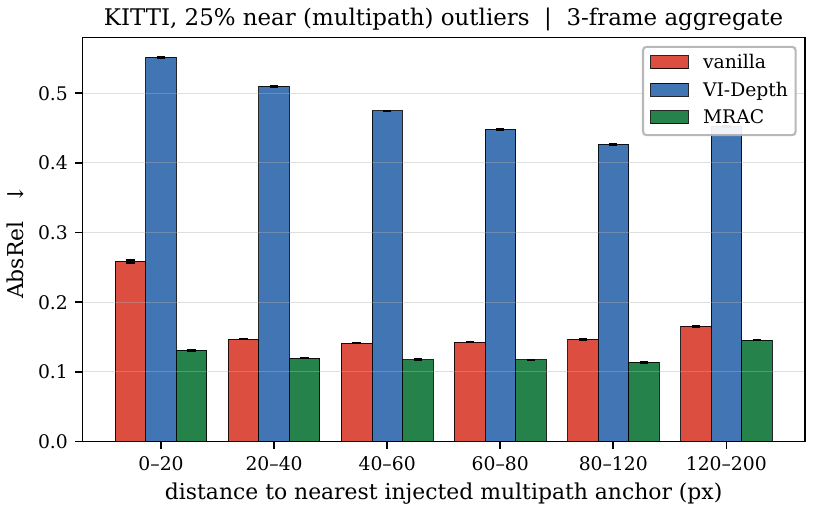}
\caption{Mechanism of the multipath blind spot, measured. For each valid
pixel in the three KITTI frames of main-paper Fig.~5, we bin by
Euclidean distance to the nearest injected outlier anchor and plot the
per-bin AbsRel. VI-Depth's error is elevated at every distance and
decays monotonically with range from the trigger---the spatial signature
of the anchor-conditioned scale-map absorption. Vanilla shows the same
shape at half the amplitude. MRAC keeps AbsRel flat at $\approx 0.12$
because its foundation-consistency gate---running at harness-wide
precision/recall $0.60/0.64$ on this cell (KITTI near, $25\%$,
$\kappa\!=\!2$; Table~\ref{tab:kappa_gate})---rejects most of the
triggering anchors before the head is called. This is the spatial
expression of the mean-P/R measurement in main-paper \S IX-B.}
\label{fig:supp:mechanism}
\end{figure*}

Section IX-A of the main paper asserts that VI-Depth's anchor-conditioned
scale map absorbs a present-but-wrong-value anchor and propagates the
resulting error into its spatial neighborhood, while MRAC's gate removes
the anchor before it can trigger the field. Fig.~\ref{fig:supp:mechanism}
makes that claim measurable. For each of the three KITTI frames shown in
main-paper Fig.~5 (25\% near outliers, $K\!=\!50$, seed 0), we compute
the Euclidean distance from every valid pixel to the nearest injected
outlier anchor, bin pixels into six distance ranges (0--20, 20--40,
40--60, 60--80, 80--120, 120--200~px), and report the per-bin AbsRel
aggregated across the three frames.

VI-Depth is elevated at every distance and shows the propagation
signature: AbsRel decays monotonically from $0.55$ at 0--20~px to
$0.42$ at 80--120~px, then plateaus. Vanilla shows the same spatial
shape at roughly half the amplitude ($0.26 \to 0.14$). MRAC is flat
and low ($\approx 0.12$) across the full distance range, because the
Theil--Sen $+$ MAD gate---at KITTI-near harness-wide precision/recall
$0.60/0.64$ (Table~\ref{tab:kappa_gate})---rejects most of the
triggering anchors before the head is called.

\section{Complete metric tables}
\label{sec:supp:fullmetrics}
This section provides the complete metric tables promised in Sec.~IV-C of
the main paper. All values are computed on the seed-$0$ anchor realization,
the same realization on which the strict-win scorecard (Table~III, main
paper) is computed; the eight headline cells of the main paper additionally
report mean\,$\pm$\,std over seeds $\{0,1,2\}$, and the seed-$0$ values here
are consistent with those spreads. Eight metrics are reported: AbsRel,
SqRel, RMSE, RMSE-log, $\log_{10}$, and $\delta_i$ at $1.25^{\,i}$; lower is
better for the first five, higher for the last three. Six methods appear:
the five-way comparison of the main paper plus the fixed-RANSAC robust
residual of Sec.~V-A. VI-Depth runs on its public
\texttt{dpt-swin2-large-384} backbone, so rows comparing it to the DAv2
methods are not backbone-matched. The clean $p{=}0$ row is the same
evaluation in all four outlier-family tables of a dataset and is repeated
in each for readability. Bold marks the best value per metric within each
corruption fraction.

\begin{table*}[t]
\centering
\caption{NYUv2, uniform outliers: complete metrics at $K{=}50$, seed~$0$.}
\footnotesize
\setlength{\tabcolsep}{4pt}
\begin{tabular}{llcccccccc}
\toprule
$p$ & Method & AbsRel & SqRel & RMSE & RMSE$_{\log}$ & $\log_{10}$ & $\delta_1$ & $\delta_2$ & $\delta_3$ \\
\midrule
0\% & vanilla (DAv2) & 0.104 & 0.065 & 0.392 & 0.141 & 0.044 & 0.891 & 0.978 & 0.994 \\
 & B$'$ (DAv2) & 0.097 & 0.059 & \textbf{0.374} & 0.132 & 0.041 & 0.904 & \textbf{0.980} & \textbf{0.995} \\
 & fixed-RANSAC residual & 0.141 & 0.200 & 0.558 & 0.195 & 0.060 & 0.839 & 0.939 & 0.971 \\
 & MRAC/van (ours) & 0.111 & 0.110 & 0.435 & 0.162 & 0.047 & 0.892 & 0.966 & 0.984 \\
 & MRAC/B$'$ (ours) & 0.100 & 0.068 & 0.415 & 0.150 & 0.045 & 0.898 & 0.970 & 0.988 \\
 & VI-Depth (swin2) & \textbf{0.079} & \textbf{0.050} & 0.390 & \textbf{0.130} & \textbf{0.036} & \textbf{0.925} & \textbf{0.981} & 0.994 \\
\midrule
10\% & vanilla (DAv2) & 0.134 & 0.104 & 0.455 & 0.165 & 0.054 & 0.839 & 0.966 & 0.993 \\
\bottomrule
\end{tabular}
\end{table*}

\begin{table*}[t]
\centering
\caption{NYUv2, near (multipath / specular) outliers: complete metrics at $K{=}50$, seed~$0$.}
\footnotesize
\setlength{\tabcolsep}{4pt}
\begin{tabular}{llcccccccc}
\toprule
$p$ & Method & AbsRel & SqRel & RMSE & RMSE$_{\log}$ & $\log_{10}$ & $\delta_1$ & $\delta_2$ & $\delta_3$ \\
\midrule
0\% & vanilla (DAv2) & 0.104 & 0.065 & 0.392 & 0.141 & 0.044 & 0.891 & 0.978 & 0.994 \\
 & B$'$ (DAv2) & 0.097 & 0.059 & \textbf{0.374} & 0.132 & 0.041 & 0.904 & \textbf{0.980} & \textbf{0.995} \\
 & fixed-RANSAC residual & 0.141 & 0.200 & 0.558 & 0.195 & 0.060 & 0.839 & 0.939 & 0.971 \\
 & MRAC/van (ours) & 0.111 & 0.110 & 0.435 & 0.162 & 0.047 & 0.892 & 0.966 & 0.984 \\
 & MRAC/B$'$ (ours) & 0.100 & 0.068 & 0.415 & 0.150 & 0.045 & 0.898 & 0.970 & 0.988 \\
 & VI-Depth (swin2) & \textbf{0.079} & \textbf{0.050} & 0.390 & \textbf{0.130} & \textbf{0.036} & \textbf{0.925} & \textbf{0.981} & 0.994 \\
\midrule
10\% & vanilla (DAv2) & 0.123 & 0.080 & 0.489 & 0.177 & 0.056 & 0.869 & 0.973 & \textbf{0.992} \\
\bottomrule
\end{tabular}
\end{table*}

\begin{table*}[t]
\centering
\caption{NYUv2, dropout (max-range / no-return) outliers: complete metrics at $K{=}50$, seed~$0$.}
\footnotesize
\setlength{\tabcolsep}{4pt}
\begin{tabular}{llcccccccc}
\toprule
$p$ & Method & AbsRel & SqRel & RMSE & RMSE$_{\log}$ & $\log_{10}$ & $\delta_1$ & $\delta_2$ & $\delta_3$ \\
\midrule
0\% & vanilla (DAv2) & 0.104 & 0.065 & 0.392 & 0.141 & 0.044 & 0.891 & 0.978 & 0.994 \\
 & B$'$ (DAv2) & 0.097 & 0.059 & \textbf{0.374} & 0.132 & 0.041 & 0.904 & \textbf{0.980} & \textbf{0.995} \\
 & fixed-RANSAC residual & 0.141 & 0.200 & 0.558 & 0.195 & 0.060 & 0.839 & 0.939 & 0.971 \\
 & MRAC/van (ours) & 0.111 & 0.110 & 0.435 & 0.162 & 0.047 & 0.892 & 0.966 & 0.984 \\
 & MRAC/B$'$ (ours) & 0.100 & 0.068 & 0.415 & 0.150 & 0.045 & 0.898 & 0.970 & 0.988 \\
 & VI-Depth (swin2) & \textbf{0.079} & \textbf{0.050} & 0.390 & \textbf{0.130} & \textbf{0.036} & \textbf{0.925} & \textbf{0.981} & 0.994 \\
\midrule
10\% & vanilla (DAv2) & 0.189 & 0.203 & 0.556 & 0.207 & 0.070 & 0.746 & 0.937 & 0.982 \\
\bottomrule
\end{tabular}
\end{table*}

\begin{table*}[t]
\centering
\caption{NYUv2, mixed-pixel outliers: complete metrics at $K{=}50$, seed~$0$.}
\footnotesize
\setlength{\tabcolsep}{4pt}
\begin{tabular}{llcccccccc}
\toprule
$p$ & Method & AbsRel & SqRel & RMSE & RMSE$_{\log}$ & $\log_{10}$ & $\delta_1$ & $\delta_2$ & $\delta_3$ \\
\midrule
0\% & vanilla (DAv2) & 0.104 & 0.065 & 0.392 & 0.141 & 0.044 & 0.891 & 0.978 & 0.994 \\
 & B$'$ (DAv2) & 0.097 & 0.059 & \textbf{0.374} & 0.132 & 0.041 & 0.904 & \textbf{0.980} & \textbf{0.995} \\
 & fixed-RANSAC residual & 0.141 & 0.200 & 0.558 & 0.195 & 0.060 & 0.839 & 0.939 & 0.971 \\
 & MRAC/van (ours) & 0.111 & 0.110 & 0.435 & 0.162 & 0.047 & 0.892 & 0.966 & 0.984 \\
 & MRAC/B$'$ (ours) & 0.100 & 0.068 & 0.415 & 0.150 & 0.045 & 0.898 & 0.970 & 0.988 \\
 & VI-Depth (swin2) & \textbf{0.079} & \textbf{0.050} & 0.390 & \textbf{0.130} & \textbf{0.036} & \textbf{0.925} & \textbf{0.981} & 0.994 \\
\midrule
10\% & vanilla (DAv2) & 0.119 & 0.080 & 0.431 & 0.154 & 0.049 & 0.866 & 0.974 & \textbf{0.993} \\
\bottomrule
\end{tabular}
\end{table*}

\begin{table*}[t]
\centering
\caption{KITTI, uniform outliers: complete metrics at $K{=}50$, seed~$0$.}
\footnotesize
\setlength{\tabcolsep}{4pt}
\begin{tabular}{llcccccccc}
\toprule
$p$ & Method & AbsRel & SqRel & RMSE & RMSE$_{\log}$ & $\log_{10}$ & $\delta_1$ & $\delta_2$ & $\delta_3$ \\
\midrule
0\% & vanilla (DAv2) & 0.133 & \textbf{0.764} & 5.175 & \textbf{0.178} & \textbf{0.056} & \textbf{0.845} & \textbf{0.967} & \textbf{0.991} \\
 & B$'$ (DAv2) & 0.146 & 0.849 & 5.351 & 0.189 & 0.061 & 0.817 & 0.963 & 0.990 \\
 & fixed-RANSAC residual & 0.192 & 2.399 & 9.626 & 0.367 & 0.107 & 0.653 & 0.821 & 0.902 \\
 & MRAC/van (ours) & \textbf{0.126} & 1.240 & 7.289 & 0.233 & 0.063 & 0.820 & 0.923 & 0.965 \\
 & MRAC/B$'$ (ours) & 0.128 & 1.263 & 7.352 & 0.236 & 0.064 & 0.822 & 0.923 & 0.963 \\
 & VI-Depth (swin2) & 0.169 & 0.807 & \textbf{5.151} & 0.228 & 0.083 & 0.710 & 0.956 & 0.989 \\
\midrule
10\% & vanilla (DAv2) & 0.361 & 2.456 & 6.591 & 0.339 & 0.129 & 0.377 & 0.817 & 0.962 \\
\bottomrule
\end{tabular}
\end{table*}

\begin{table*}[t]
\centering
\caption{KITTI, near (multipath / specular) outliers: complete metrics at $K{=}50$, seed~$0$.}
\footnotesize
\setlength{\tabcolsep}{4pt}
\begin{tabular}{llcccccccc}
\toprule
$p$ & Method & AbsRel & SqRel & RMSE & RMSE$_{\log}$ & $\log_{10}$ & $\delta_1$ & $\delta_2$ & $\delta_3$ \\
\midrule
0\% & vanilla (DAv2) & 0.133 & \textbf{0.764} & 5.175 & \textbf{0.178} & \textbf{0.056} & \textbf{0.845} & \textbf{0.967} & \textbf{0.991} \\
 & B$'$ (DAv2) & 0.146 & 0.849 & 5.351 & 0.189 & 0.061 & 0.817 & 0.963 & 0.990 \\
 & fixed-RANSAC residual & 0.192 & 2.399 & 9.626 & 0.367 & 0.107 & 0.653 & 0.821 & 0.902 \\
 & MRAC/van (ours) & \textbf{0.126} & 1.240 & 7.289 & 0.233 & 0.063 & 0.820 & 0.923 & 0.965 \\
 & MRAC/B$'$ (ours) & 0.128 & 1.263 & 7.352 & 0.236 & 0.064 & 0.822 & 0.923 & 0.963 \\
 & VI-Depth (swin2) & 0.169 & 0.807 & \textbf{5.151} & 0.228 & 0.083 & 0.710 & 0.956 & 0.989 \\
\midrule
10\% & vanilla (DAv2) & \textbf{0.122} & \textbf{0.795} & \textbf{5.619} & \textbf{0.187} & \textbf{0.056} & \textbf{0.853} & \textbf{0.959} & \textbf{0.986} \\
\bottomrule
\end{tabular}
\end{table*}

\begin{table*}[t]
\centering
\caption{KITTI, dropout (max-range / no-return) outliers: complete metrics at $K{=}50$, seed~$0$.}
\footnotesize
\setlength{\tabcolsep}{4pt}
\begin{tabular}{llcccccccc}
\toprule
$p$ & Method & AbsRel & SqRel & RMSE & RMSE$_{\log}$ & $\log_{10}$ & $\delta_1$ & $\delta_2$ & $\delta_3$ \\
\midrule
0\% & vanilla (DAv2) & 0.133 & \textbf{0.764} & 5.175 & \textbf{0.178} & \textbf{0.056} & \textbf{0.845} & \textbf{0.967} & \textbf{0.991} \\
 & B$'$ (DAv2) & 0.146 & 0.849 & 5.351 & 0.189 & 0.061 & 0.817 & 0.963 & 0.990 \\
 & fixed-RANSAC residual & 0.192 & 2.399 & 9.626 & 0.367 & 0.107 & 0.653 & 0.821 & 0.902 \\
 & MRAC/van (ours) & \textbf{0.126} & 1.240 & 7.289 & 0.233 & 0.063 & 0.820 & 0.923 & 0.965 \\
 & MRAC/B$'$ (ours) & 0.128 & 1.263 & 7.352 & 0.236 & 0.064 & 0.822 & 0.923 & 0.963 \\
 & VI-Depth (swin2) & 0.169 & 0.807 & \textbf{5.151} & 0.228 & 0.083 & 0.710 & 0.956 & 0.989 \\
\midrule
10\% & vanilla (DAv2) & 0.683 & 6.735 & 9.124 & 0.541 & 0.213 & 0.151 & 0.423 & 0.796 \\
\bottomrule
\end{tabular}
\end{table*}

\begin{table*}[t]
\centering
\caption{KITTI, mixed-pixel outliers: complete metrics at $K{=}50$, seed~$0$.}
\footnotesize
\setlength{\tabcolsep}{4pt}
\begin{tabular}{llcccccccc}
\toprule
$p$ & Method & AbsRel & SqRel & RMSE & RMSE$_{\log}$ & $\log_{10}$ & $\delta_1$ & $\delta_2$ & $\delta_3$ \\
\midrule
0\% & vanilla (DAv2) & 0.133 & \textbf{0.764} & 5.175 & \textbf{0.178} & \textbf{0.056} & \textbf{0.845} & \textbf{0.967} & \textbf{0.991} \\
 & B$'$ (DAv2) & 0.146 & 0.849 & 5.351 & 0.189 & 0.061 & 0.817 & 0.963 & 0.990 \\
 & fixed-RANSAC residual & 0.192 & 2.399 & 9.626 & 0.367 & 0.107 & 0.653 & 0.821 & 0.902 \\
 & MRAC/van (ours) & \textbf{0.126} & 1.240 & 7.289 & 0.233 & 0.063 & 0.820 & 0.923 & 0.965 \\
 & MRAC/B$'$ (ours) & 0.128 & 1.263 & 7.352 & 0.236 & 0.064 & 0.822 & 0.923 & 0.963 \\
 & VI-Depth (swin2) & 0.169 & 0.807 & \textbf{5.151} & 0.228 & 0.083 & 0.710 & 0.956 & 0.989 \\
\midrule
10\% & vanilla (DAv2) & 0.159 & \textbf{0.984} & \textbf{5.674} & \textbf{0.204} & 0.066 & 0.782 & \textbf{0.955} & \textbf{0.987} \\
\bottomrule
\end{tabular}
\end{table*}

\begin{table*}[t]
\centering
\caption{DIODE, uniform outliers: complete metrics at $K{=}50$, seed~$0$.}
\footnotesize
\setlength{\tabcolsep}{4pt}
\begin{tabular}{llcccccccc}
\toprule
$p$ & Method & AbsRel & SqRel & RMSE & RMSE$_{\log}$ & $\log_{10}$ & $\delta_1$ & $\delta_2$ & $\delta_3$ \\
\midrule
0\% & vanilla (DAv2) & 0.351 & 3.944 & 3.705 & 0.318 & 0.092 & 0.734 & 0.873 & 0.928 \\
 & B$'$ (DAv2) & 0.348 & 3.921 & \textbf{3.701} & 0.315 & 0.092 & 0.736 & 0.875 & 0.929 \\
 & fixed-RANSAC residual & 0.322 & 5.278 & 4.923 & 0.376 & 0.105 & 0.738 & 0.845 & 0.899 \\
 & MRAC/van (ours) & 0.314 & 4.284 & 4.148 & 0.328 & 0.088 & 0.769 & 0.878 & 0.926 \\
 & MRAC/B$'$ (ours) & 0.304 & 4.253 & 4.132 & 0.320 & \textbf{0.085} & \textbf{0.779} & \textbf{0.883} & 0.928 \\
 & VI-Depth (swin2) & \textbf{0.218} & \textbf{2.129} & 4.180 & \textbf{0.295} & 0.088 & 0.743 & 0.873 & \textbf{0.930} \\
\midrule
10\% & vanilla (DAv2) & 0.530 & 5.695 & 4.337 & 0.417 & 0.134 & 0.588 & 0.773 & 0.872 \\
\bottomrule
\end{tabular}
\end{table*}

\begin{table*}[t]
\centering
\caption{DIODE, near (multipath / specular) outliers: complete metrics at $K{=}50$, seed~$0$.}
\footnotesize
\setlength{\tabcolsep}{4pt}
\begin{tabular}{llcccccccc}
\toprule
$p$ & Method & AbsRel & SqRel & RMSE & RMSE$_{\log}$ & $\log_{10}$ & $\delta_1$ & $\delta_2$ & $\delta_3$ \\
\midrule
0\% & vanilla (DAv2) & 0.351 & 3.944 & 3.705 & 0.318 & 0.092 & 0.734 & 0.873 & 0.928 \\
 & B$'$ (DAv2) & 0.348 & 3.921 & \textbf{3.701} & 0.315 & 0.092 & 0.736 & 0.875 & 0.929 \\
 & fixed-RANSAC residual & 0.322 & 5.278 & 4.923 & 0.376 & 0.105 & 0.738 & 0.845 & 0.899 \\
 & MRAC/van (ours) & 0.314 & 4.284 & 4.148 & 0.328 & 0.088 & 0.769 & 0.878 & 0.926 \\
 & MRAC/B$'$ (ours) & 0.304 & 4.253 & 4.132 & 0.320 & \textbf{0.085} & \textbf{0.779} & \textbf{0.883} & 0.928 \\
 & VI-Depth (swin2) & \textbf{0.218} & \textbf{2.129} & 4.180 & \textbf{0.295} & 0.088 & 0.743 & 0.873 & \textbf{0.930} \\
\midrule
10\% & vanilla (DAv2) & 0.345 & 3.481 & 3.889 & 0.331 & 0.101 & 0.718 & 0.875 & \textbf{0.931} \\
\bottomrule
\end{tabular}
\end{table*}

\begin{table*}[t]
\centering
\caption{DIODE, dropout (max-range / no-return) outliers: complete metrics at $K{=}50$, seed~$0$.}
\footnotesize
\setlength{\tabcolsep}{4pt}
\begin{tabular}{llcccccccc}
\toprule
$p$ & Method & AbsRel & SqRel & RMSE & RMSE$_{\log}$ & $\log_{10}$ & $\delta_1$ & $\delta_2$ & $\delta_3$ \\
\midrule
0\% & vanilla (DAv2) & 0.351 & 3.944 & 3.705 & 0.318 & 0.092 & 0.734 & 0.873 & 0.928 \\
 & B$'$ (DAv2) & 0.348 & 3.921 & \textbf{3.701} & 0.315 & 0.092 & 0.736 & 0.875 & 0.929 \\
 & fixed-RANSAC residual & 0.322 & 5.278 & 4.923 & 0.376 & 0.105 & 0.738 & 0.845 & 0.899 \\
 & MRAC/van (ours) & 0.314 & 4.284 & 4.148 & 0.328 & 0.088 & 0.769 & 0.878 & 0.926 \\
 & MRAC/B$'$ (ours) & 0.304 & 4.253 & 4.132 & 0.320 & \textbf{0.085} & \textbf{0.779} & \textbf{0.883} & 0.928 \\
 & VI-Depth (swin2) & \textbf{0.218} & \textbf{2.129} & 4.180 & \textbf{0.295} & 0.088 & 0.743 & 0.873 & \textbf{0.930} \\
\midrule
10\% & vanilla (DAv2) & 0.815 & 9.772 & 5.510 & 0.549 & 0.189 & 0.435 & 0.655 & 0.770 \\
\bottomrule
\end{tabular}
\end{table*}

\begin{table*}[t]
\centering
\caption{DIODE, mixed-pixel outliers: complete metrics at $K{=}50$, seed~$0$.}
\footnotesize
\setlength{\tabcolsep}{4pt}
\begin{tabular}{llcccccccc}
\toprule
$p$ & Method & AbsRel & SqRel & RMSE & RMSE$_{\log}$ & $\log_{10}$ & $\delta_1$ & $\delta_2$ & $\delta_3$ \\
\midrule
0\% & vanilla (DAv2) & 0.351 & 3.944 & 3.705 & 0.318 & 0.092 & 0.734 & 0.873 & 0.928 \\
 & B$'$ (DAv2) & 0.348 & 3.921 & \textbf{3.701} & 0.315 & 0.092 & 0.736 & 0.875 & 0.929 \\
 & fixed-RANSAC residual & 0.322 & 5.278 & 4.923 & 0.376 & 0.105 & 0.738 & 0.845 & 0.899 \\
 & MRAC/van (ours) & 0.314 & 4.284 & 4.148 & 0.328 & 0.088 & 0.769 & 0.878 & 0.926 \\
 & MRAC/B$'$ (ours) & 0.304 & 4.253 & 4.132 & 0.320 & \textbf{0.085} & \textbf{0.779} & \textbf{0.883} & 0.928 \\
 & VI-Depth (swin2) & \textbf{0.218} & \textbf{2.129} & 4.180 & \textbf{0.295} & 0.088 & 0.743 & 0.873 & \textbf{0.930} \\
\midrule
10\% & vanilla (DAv2) & 0.367 & 4.024 & 3.832 & 0.328 & 0.098 & 0.713 & 0.863 & 0.923 \\
\bottomrule
\end{tabular}
\end{table*}

\begin{table*}[t]
\centering
\caption{SUN RGB-D, uniform outliers: complete metrics at $K{=}50$, seed~$0$.}
\footnotesize
\setlength{\tabcolsep}{4pt}
\begin{tabular}{llcccccccc}
\toprule
$p$ & Method & AbsRel & SqRel & RMSE & RMSE$_{\log}$ & $\log_{10}$ & $\delta_1$ & $\delta_2$ & $\delta_3$ \\
\midrule
0\% & vanilla (DAv2) & 0.149 & 0.219 & 0.478 & 0.196 & 0.053 & 0.875 & 0.956 & 0.976 \\
 & B$'$ (DAv2) & 0.142 & 0.214 & 0.458 & 0.186 & 0.050 & 0.888 & \textbf{0.960} & 0.978 \\
 & fixed-RANSAC residual & 0.174 & 0.315 & 0.580 & 0.225 & 0.063 & 0.845 & 0.936 & 0.965 \\
 & MRAC/van (ours) & 0.163 & 0.269 & 0.519 & 0.213 & 0.057 & 0.874 & 0.951 & 0.971 \\
 & MRAC/B$'$ (ours) & 0.147 & 0.235 & 0.492 & 0.199 & 0.052 & 0.886 & 0.955 & 0.973 \\
 & VI-Depth (swin2) & \textbf{0.114} & \textbf{0.143} & \textbf{0.448} & \textbf{0.169} & \textbf{0.045} & \textbf{0.891} & 0.959 & \textbf{0.980} \\
\midrule
10\% & vanilla (DAv2) & 0.181 & 0.257 & 0.527 & 0.218 & 0.063 & 0.827 & 0.947 & 0.974 \\
\bottomrule
\end{tabular}
\end{table*}

\begin{table*}[t]
\centering
\caption{SUN RGB-D, near (multipath / specular) outliers: complete metrics at $K{=}50$, seed~$0$.}
\footnotesize
\setlength{\tabcolsep}{4pt}
\begin{tabular}{llcccccccc}
\toprule
$p$ & Method & AbsRel & SqRel & RMSE & RMSE$_{\log}$ & $\log_{10}$ & $\delta_1$ & $\delta_2$ & $\delta_3$ \\
\midrule
0\% & vanilla (DAv2) & 0.149 & 0.219 & 0.478 & 0.196 & 0.053 & 0.875 & 0.956 & 0.976 \\
 & B$'$ (DAv2) & 0.142 & 0.214 & 0.458 & 0.186 & 0.050 & 0.888 & \textbf{0.960} & 0.978 \\
 & fixed-RANSAC residual & 0.174 & 0.315 & 0.580 & 0.225 & 0.063 & 0.845 & 0.936 & 0.965 \\
 & MRAC/van (ours) & 0.163 & 0.269 & 0.519 & 0.213 & 0.057 & 0.874 & 0.951 & 0.971 \\
 & MRAC/B$'$ (ours) & 0.147 & 0.235 & 0.492 & 0.199 & 0.052 & 0.886 & 0.955 & 0.973 \\
 & VI-Depth (swin2) & \textbf{0.114} & \textbf{0.143} & \textbf{0.448} & \textbf{0.169} & \textbf{0.045} & \textbf{0.891} & 0.959 & \textbf{0.980} \\
\midrule
10\% & vanilla (DAv2) & 0.174 & 0.213 & 0.550 & 0.232 & 0.070 & 0.828 & 0.948 & 0.974 \\
\bottomrule
\end{tabular}
\end{table*}

\begin{table*}[t]
\centering
\caption{SUN RGB-D, dropout (max-range / no-return) outliers: complete metrics at $K{=}50$, seed~$0$.}
\footnotesize
\setlength{\tabcolsep}{4pt}
\begin{tabular}{llcccccccc}
\toprule
$p$ & Method & AbsRel & SqRel & RMSE & RMSE$_{\log}$ & $\log_{10}$ & $\delta_1$ & $\delta_2$ & $\delta_3$ \\
\midrule
0\% & vanilla (DAv2) & 0.149 & 0.219 & 0.478 & 0.196 & 0.053 & 0.875 & 0.956 & 0.976 \\
 & B$'$ (DAv2) & 0.142 & 0.214 & 0.458 & 0.186 & 0.050 & 0.888 & \textbf{0.960} & 0.978 \\
 & fixed-RANSAC residual & 0.174 & 0.315 & 0.580 & 0.225 & 0.063 & 0.845 & 0.936 & 0.965 \\
 & MRAC/van (ours) & 0.163 & 0.269 & 0.519 & 0.213 & 0.057 & 0.874 & 0.951 & 0.971 \\
 & MRAC/B$'$ (ours) & 0.147 & 0.235 & 0.492 & 0.199 & 0.052 & 0.886 & 0.955 & 0.973 \\
 & VI-Depth (swin2) & \textbf{0.114} & \textbf{0.143} & \textbf{0.448} & \textbf{0.169} & \textbf{0.045} & \textbf{0.891} & 0.959 & \textbf{0.980} \\
\midrule
10\% & vanilla (DAv2) & 0.247 & 0.390 & 0.638 & 0.264 & 0.081 & 0.725 & 0.918 & 0.964 \\
\bottomrule
\end{tabular}
\end{table*}

\begin{table*}[t]
\centering
\caption{SUN RGB-D, mixed-pixel outliers: complete metrics at $K{=}50$, seed~$0$.}
\footnotesize
\setlength{\tabcolsep}{4pt}
\begin{tabular}{llcccccccc}
\toprule
$p$ & Method & AbsRel & SqRel & RMSE & RMSE$_{\log}$ & $\log_{10}$ & $\delta_1$ & $\delta_2$ & $\delta_3$ \\
\midrule
0\% & vanilla (DAv2) & 0.149 & 0.219 & 0.478 & 0.196 & 0.053 & 0.875 & 0.956 & 0.976 \\
 & B$'$ (DAv2) & 0.142 & 0.214 & 0.458 & 0.186 & 0.050 & 0.888 & \textbf{0.960} & 0.978 \\
 & fixed-RANSAC residual & 0.174 & 0.315 & 0.580 & 0.225 & 0.063 & 0.845 & 0.936 & 0.965 \\
 & MRAC/van (ours) & 0.163 & 0.269 & 0.519 & 0.213 & 0.057 & 0.874 & 0.951 & 0.971 \\
 & MRAC/B$'$ (ours) & 0.147 & 0.235 & 0.492 & 0.199 & 0.052 & 0.886 & 0.955 & 0.973 \\
 & VI-Depth (swin2) & \textbf{0.114} & \textbf{0.143} & \textbf{0.448} & \textbf{0.169} & \textbf{0.045} & \textbf{0.891} & 0.959 & \textbf{0.980} \\
\midrule
10\% & vanilla (DAv2) & 0.161 & 0.220 & 0.500 & 0.205 & 0.058 & 0.855 & 0.952 & 0.975 \\
\bottomrule
\end{tabular}
\end{table*}

\section{Closed-form CFA references}
\label{sec:supp:cfa}
AbsRel of the three closed-form CFA references (ordinary least squares,
Huber, RANSAC) that anchor absolute error levels in Sec.~V of the main
paper, across the full grid ($K{=}50$, seed~$0$). Bold: best per row.

\begin{table}[t]
\centering
\caption{NYUv2: CFA references, AbsRel across the outlier grid.}
\footnotesize
\setlength{\tabcolsep}{5pt}
\begin{tabular}{llccc}
\toprule
Family & $p$ & cfa-ols & cfa-huber & cfa-ransac \\
\midrule
uniform & 0\% & 0.155 & \textbf{0.144} & 0.155 \\
uniform & 10\% & 0.169 & \textbf{0.144} & 0.156 \\
uniform & 25\% & 0.204 & \textbf{0.153} & 0.158 \\
uniform & 40\% & 0.261 & 0.175 & \textbf{0.164} \\
\midrule
near & 0\% & 0.155 & \textbf{0.144} & 0.155 \\
near & 10\% & 0.155 & \textbf{0.141} & 0.155 \\
near & 25\% & 0.201 & \textbf{0.152} & 0.164 \\
near & 40\% & 0.297 & 0.219 & \textbf{0.193} \\
\midrule
dropout & 0\% & 0.155 & \textbf{0.144} & 0.155 \\
dropout & 10\% & 0.216 & \textbf{0.153} & 0.156 \\
dropout & 25\% & 0.352 & 0.188 & \textbf{0.171} \\
dropout & 40\% & 0.534 & \textbf{0.380} & 0.774 \\
\midrule
mixed-pixel & 0\% & 0.155 & \textbf{0.144} & 0.155 \\
mixed-pixel & 10\% & 0.158 & \textbf{0.143} & 0.155 \\
mixed-pixel & 25\% & 0.171 & \textbf{0.146} & 0.158 \\
mixed-pixel & 40\% & 0.197 & \textbf{0.157} & 0.165 \\
\bottomrule
\end{tabular}
\end{table}

\begin{table}[t]
\centering
\caption{KITTI: CFA references, AbsRel across the outlier grid.}
\footnotesize
\setlength{\tabcolsep}{5pt}
\begin{tabular}{llccc}
\toprule
Family & $p$ & cfa-ols & cfa-huber & cfa-ransac \\
\midrule
uniform & 0\% & 0.333 & 0.233 & \textbf{0.204} \\
uniform & 10\% & 0.440 & 0.246 & \textbf{0.203} \\
uniform & 25\% & 0.722 & 0.287 & \textbf{0.207} \\
uniform & 40\% & 1.133 & 0.457 & \textbf{0.214} \\
\midrule
near & 0\% & 0.333 & 0.233 & \textbf{0.204} \\
near & 10\% & 0.297 & 0.221 & \textbf{0.207} \\
near & 25\% & 0.276 & 0.215 & \textbf{0.212} \\
near & 40\% & 0.315 & \textbf{0.239} & 0.264 \\
\midrule
dropout & 0\% & 0.333 & 0.233 & \textbf{0.204} \\
dropout & 10\% & 0.701 & 0.264 & \textbf{0.203} \\
dropout & 25\% & 1.481 & \textbf{0.374} & 2.921 \\
dropout & 40\% & 2.417 & \textbf{1.155} & 5.955 \\
\midrule
mixed-pixel & 0\% & 0.333 & 0.233 & \textbf{0.204} \\
mixed-pixel & 10\% & 0.311 & 0.221 & \textbf{0.206} \\
mixed-pixel & 25\% & 0.301 & \textbf{0.208} & 0.211 \\
mixed-pixel & 40\% & 0.322 & \textbf{0.205} & 0.234 \\
\bottomrule
\end{tabular}
\end{table}

\begin{table}[t]
\centering
\caption{DIODE: CFA references, AbsRel across the outlier grid.}
\footnotesize
\setlength{\tabcolsep}{5pt}
\begin{tabular}{llccc}
\toprule
Family & $p$ & cfa-ols & cfa-huber & cfa-ransac \\
\midrule
uniform & 0\% & 0.386 & 0.339 & \textbf{0.321} \\
uniform & 10\% & 0.511 & 0.358 & \textbf{0.320} \\
uniform & 25\% & 0.744 & 0.404 & \textbf{0.338} \\
uniform & 40\% & 1.040 & 0.538 & \textbf{0.344} \\
\midrule
near & 0\% & 0.386 & 0.339 & \textbf{0.321} \\
near & 10\% & 0.375 & \textbf{0.317} & 0.320 \\
near & 25\% & 0.392 & \textbf{0.302} & 0.313 \\
near & 40\% & 0.437 & 0.334 & \textbf{0.330} \\
\midrule
dropout & 0\% & 0.386 & 0.339 & \textbf{0.321} \\
dropout & 10\% & 0.753 & 0.389 & \textbf{0.360} \\
dropout & 25\% & 1.403 & \textbf{0.509} & 1.821 \\
dropout & 40\% & 2.190 & \textbf{1.224} & 3.545 \\
\midrule
mixed-pixel & 0\% & 0.386 & 0.339 & \textbf{0.321} \\
mixed-pixel & 10\% & 0.384 & 0.334 & \textbf{0.314} \\
mixed-pixel & 25\% & 0.384 & \textbf{0.325} & 0.329 \\
mixed-pixel & 40\% & 0.412 & \textbf{0.329} & 0.334 \\
\bottomrule
\end{tabular}
\end{table}

\begin{table}[t]
\centering
\caption{SUN RGB-D: CFA references, AbsRel across the outlier grid.}
\footnotesize
\setlength{\tabcolsep}{5pt}
\begin{tabular}{llccc}
\toprule
Family & $p$ & cfa-ols & cfa-huber & cfa-ransac \\
\midrule
uniform & 0\% & 0.173 & \textbf{0.166} & \textbf{0.167} \\
uniform & 10\% & 0.193 & 0.169 & \textbf{0.167} \\
uniform & 25\% & 0.240 & 0.177 & \textbf{0.168} \\
uniform & 40\% & 0.312 & 0.204 & \textbf{0.175} \\
\midrule
near & 0\% & 0.173 & \textbf{0.166} & \textbf{0.167} \\
near & 10\% & 0.176 & \textbf{0.163} & 0.169 \\
near & 25\% & 0.232 & 0.178 & \textbf{0.172} \\
near & 40\% & 0.325 & 0.254 & \textbf{0.198} \\
\midrule
dropout & 0\% & 0.173 & \textbf{0.166} & \textbf{0.167} \\
dropout & 10\% & 0.254 & 0.180 & \textbf{0.168} \\
dropout & 25\% & 0.425 & 0.223 & \textbf{0.181} \\
dropout & 40\% & 0.645 & \textbf{0.422} & 0.674 \\
\midrule
mixed-pixel & 0\% & 0.173 & \textbf{0.166} & \textbf{0.167} \\
mixed-pixel & 10\% & 0.173 & \textbf{0.164} & 0.167 \\
mixed-pixel & 25\% & 0.182 & \textbf{0.163} & 0.168 \\
mixed-pixel & 40\% & 0.208 & \textbf{0.172} & 0.173 \\
\bottomrule
\end{tabular}
\end{table}

\section{VI-Depth at its training budget ($K{=}150$)}
\label{sec:supp:k150}
Complete metric grid for VI-Depth official evaluated at its training anchor
budget $K{=}150$ (seed~$0$), expanding the near/$25\%$ AbsRel values quoted
in Sec.~IX-B of the main paper.

\begin{table*}[t]
\centering
\caption{NYUv2: VI-Depth official at $K{=}150$, complete metrics (seed~$0$).}
\footnotesize
\setlength{\tabcolsep}{4pt}
\begin{tabular}{llcccccccc}
\toprule
Family & $p$ & AbsRel & SqRel & RMSE & RMSE$_{\log}$ & $\log_{10}$ & $\delta_1$ & $\delta_2$ & $\delta_3$ \\
\midrule
uniform & 0\% & 0.063 & 0.037 & 0.334 & 0.108 & 0.029 & 0.948 & 0.987 & 0.996 \\
uniform & 10\% & 0.081 & 0.049 & 0.387 & 0.124 & 0.036 & 0.925 & 0.984 & 0.996 \\
uniform & 25\% & 0.114 & 0.069 & 0.456 & 0.150 & 0.050 & 0.879 & 0.981 & 0.996 \\
uniform & 40\% & 0.155 & 0.105 & 0.549 & 0.186 & 0.066 & 0.773 & 0.970 & 0.996 \\
\midrule
near & 0\% & 0.063 & 0.037 & 0.334 & 0.108 & 0.029 & 0.948 & 0.987 & 0.996 \\
near & 10\% & 0.145 & 0.132 & 0.679 & 0.215 & 0.072 & 0.756 & 0.944 & 0.984 \\
near & 25\% & 0.271 & 0.357 & 1.118 & 0.386 & 0.146 & 0.332 & 0.722 & 0.924 \\
near & 40\% & 0.371 & 0.593 & 1.436 & 0.551 & 0.217 & 0.180 & 0.426 & 0.735 \\
\midrule
dropout & 0\% & 0.063 & 0.037 & 0.334 & 0.108 & 0.029 & 0.948 & 0.987 & 0.996 \\
dropout & 10\% & 0.082 & 0.051 & 0.370 & 0.126 & 0.035 & 0.922 & 0.982 & 0.996 \\
dropout & 25\% & 0.150 & 0.106 & 0.485 & 0.180 & 0.058 & 0.815 & 0.965 & 0.994 \\
dropout & 40\% & 0.257 & 0.234 & 0.688 & 0.261 & 0.094 & 0.594 & 0.900 & 0.984 \\
\midrule
mixed-pixel & 0\% & 0.063 & 0.037 & 0.334 & 0.108 & 0.029 & 0.948 & 0.987 & 0.996 \\
mixed-pixel & 10\% & 0.080 & 0.052 & 0.403 & 0.126 & 0.037 & 0.922 & 0.982 & 0.995 \\
mixed-pixel & 25\% & 0.110 & 0.078 & 0.501 & 0.154 & 0.050 & 0.870 & 0.975 & 0.994 \\
mixed-pixel & 40\% & 0.143 & 0.115 & 0.608 & 0.189 & 0.065 & 0.784 & 0.960 & 0.991 \\
\bottomrule
\end{tabular}
\end{table*}

\begin{table*}[t]
\centering
\caption{KITTI: VI-Depth official at $K{=}150$, complete metrics (seed~$0$).}
\footnotesize
\setlength{\tabcolsep}{4pt}
\begin{tabular}{llcccccccc}
\toprule
Family & $p$ & AbsRel & SqRel & RMSE & RMSE$_{\log}$ & $\log_{10}$ & $\delta_1$ & $\delta_2$ & $\delta_3$ \\
\midrule
uniform & 0\% & 0.160 & 0.708 & 4.833 & 0.208 & 0.078 & 0.753 & 0.973 & 0.994 \\
uniform & 10\% & 0.127 & 0.720 & 5.207 & 0.185 & 0.061 & 0.831 & 0.967 & 0.991 \\
uniform & 25\% & 0.140 & 0.831 & 5.528 & 0.195 & 0.062 & 0.814 & 0.961 & 0.988 \\
uniform & 40\% & 0.229 & 1.291 & 6.218 & 0.262 & 0.093 & 0.591 & 0.928 & 0.983 \\
\midrule
near & 0\% & 0.160 & 0.708 & 4.833 & 0.208 & 0.078 & 0.753 & 0.973 & 0.994 \\
near & 10\% & 0.338 & 2.805 & 9.353 & 0.487 & 0.189 & 0.132 & 0.625 & 0.865 \\
near & 25\% & 0.490 & 4.756 & 11.664 & 0.738 & 0.305 & 0.008 & 0.096 & 0.521 \\
near & 40\% & 0.590 & 6.381 & 13.223 & 0.949 & 0.400 & 0.001 & 0.009 & 0.101 \\
\midrule
dropout & 0\% & 0.160 & 0.708 & 4.833 & 0.208 & 0.078 & 0.753 & 0.973 & 0.994 \\
dropout & 10\% & 0.123 & 0.576 & 4.446 & 0.172 & 0.057 & 0.845 & 0.978 & 0.995 \\
dropout & 25\% & 0.174 & 0.865 & 4.665 & 0.203 & 0.070 & 0.740 & 0.971 & 0.995 \\
dropout & 40\% & 0.355 & 2.315 & 6.354 & 0.336 & 0.127 & 0.368 & 0.812 & 0.983 \\
\midrule
mixed-pixel & 0\% & 0.160 & 0.708 & 4.833 & 0.208 & 0.078 & 0.753 & 0.973 & 0.994 \\
mixed-pixel & 10\% & 0.154 & 1.142 & 6.690 & 0.238 & 0.078 & 0.750 & 0.933 & 0.976 \\
mixed-pixel & 25\% & 0.161 & 1.608 & 8.129 & 0.285 & 0.084 & 0.725 & 0.892 & 0.950 \\
mixed-pixel & 40\% & 0.190 & 2.159 & 9.338 & 0.350 & 0.102 & 0.662 & 0.843 & 0.919 \\
\bottomrule
\end{tabular}
\end{table*}

\begin{table*}[t]
\centering
\caption{DIODE: VI-Depth official at $K{=}150$, complete metrics (seed~$0$).}
\footnotesize
\setlength{\tabcolsep}{4pt}
\begin{tabular}{llcccccccc}
\toprule
Family & $p$ & AbsRel & SqRel & RMSE & RMSE$_{\log}$ & $\log_{10}$ & $\delta_1$ & $\delta_2$ & $\delta_3$ \\
\midrule
uniform & 0\% & 0.209 & 2.017 & 4.001 & 0.275 & 0.082 & 0.757 & 0.881 & 0.936 \\
uniform & 10\% & 0.228 & 2.053 & 4.228 & 0.292 & 0.090 & 0.727 & 0.868 & 0.931 \\
uniform & 25\% & 0.277 & 2.212 & 4.520 & 0.328 & 0.109 & 0.640 & 0.846 & 0.925 \\
uniform & 40\% & 0.346 & 2.444 & 4.896 & 0.380 & 0.135 & 0.490 & 0.789 & 0.905 \\
\midrule
near & 0\% & 0.209 & 2.017 & 4.001 & 0.275 & 0.082 & 0.757 & 0.881 & 0.936 \\
near & 10\% & 0.307 & 2.488 & 5.461 & 0.413 & 0.150 & 0.408 & 0.753 & 0.884 \\
near & 25\% & 0.432 & 3.469 & 6.982 & 0.631 & 0.250 & 0.160 & 0.359 & 0.658 \\
near & 40\% & 0.523 & 4.469 & 7.960 & 0.822 & 0.335 & 0.092 & 0.200 & 0.365 \\
\midrule
dropout & 0\% & 0.209 & 2.017 & 4.001 & 0.275 & 0.082 & 0.757 & 0.881 & 0.936 \\
dropout & 10\% & 0.229 & 2.080 & 3.953 & 0.284 & 0.083 & 0.756 & 0.884 & 0.938 \\
dropout & 25\% & 0.326 & 2.580 & 4.103 & 0.337 & 0.107 & 0.627 & 0.864 & 0.934 \\
dropout & 40\% & 0.495 & 3.792 & 4.695 & 0.431 & 0.150 & 0.383 & 0.744 & 0.906 \\
\midrule
mixed-pixel & 0\% & 0.209 & 2.017 & 4.001 & 0.275 & 0.082 & 0.757 & 0.881 & 0.936 \\
mixed-pixel & 10\% & 0.214 & 2.012 & 4.262 & 0.289 & 0.087 & 0.740 & 0.869 & 0.929 \\
mixed-pixel & 25\% & 0.229 & 2.023 & 4.612 & 0.315 & 0.097 & 0.712 & 0.850 & 0.917 \\
mixed-pixel & 40\% & 0.252 & 2.200 & 5.040 & 0.350 & 0.111 & 0.673 & 0.825 & 0.900 \\
\bottomrule
\end{tabular}
\end{table*}

\begin{table*}[t]
\centering
\caption{SUN RGB-D: VI-Depth official at $K{=}150$, complete metrics (seed~$0$).}
\footnotesize
\setlength{\tabcolsep}{4pt}
\begin{tabular}{llcccccccc}
\toprule
Family & $p$ & AbsRel & SqRel & RMSE & RMSE$_{\log}$ & $\log_{10}$ & $\delta_1$ & $\delta_2$ & $\delta_3$ \\
\midrule
uniform & 0\% & 0.096 & 0.110 & 0.395 & 0.147 & 0.039 & 0.903 & 0.965 & 0.984 \\
uniform & 10\% & 0.114 & 0.121 & 0.439 & 0.163 & 0.047 & 0.881 & 0.959 & 0.981 \\
uniform & 25\% & 0.150 & 0.143 & 0.502 & 0.190 & 0.061 & 0.832 & 0.952 & 0.979 \\
uniform & 40\% & 0.190 & 0.174 & 0.568 & 0.222 & 0.076 & 0.742 & 0.941 & 0.976 \\
\midrule
near & 0\% & 0.096 & 0.110 & 0.395 & 0.147 & 0.039 & 0.903 & 0.965 & 0.984 \\
near & 10\% & 0.161 & 0.170 & 0.643 & 0.229 & 0.075 & 0.753 & 0.921 & 0.971 \\
near & 25\% & 0.258 & 0.315 & 0.970 & 0.365 & 0.135 & 0.419 & 0.759 & 0.922 \\
near & 40\% & 0.341 & 0.484 & 1.215 & 0.500 & 0.193 & 0.245 & 0.538 & 0.792 \\
\midrule
dropout & 0\% & 0.096 & 0.110 & 0.395 & 0.147 & 0.039 & 0.903 & 0.965 & 0.984 \\
dropout & 10\% & 0.122 & 0.136 & 0.439 & 0.170 & 0.047 & 0.881 & 0.959 & 0.981 \\
dropout & 25\% & 0.202 & 0.212 & 0.547 & 0.230 & 0.073 & 0.763 & 0.939 & 0.976 \\
dropout & 40\% & 0.322 & 0.372 & 0.729 & 0.312 & 0.110 & 0.524 & 0.872 & 0.965 \\
\midrule
mixed-pixel & 0\% & 0.096 & 0.110 & 0.395 & 0.147 & 0.039 & 0.903 & 0.965 & 0.984 \\
mixed-pixel & 10\% & 0.111 & 0.116 & 0.450 & 0.163 & 0.047 & 0.878 & 0.956 & 0.981 \\
mixed-pixel & 25\% & 0.138 & 0.136 & 0.536 & 0.190 & 0.060 & 0.831 & 0.943 & 0.976 \\
mixed-pixel & 40\% & 0.165 & 0.160 & 0.613 & 0.219 & 0.072 & 0.770 & 0.927 & 0.970 \\
\bottomrule
\end{tabular}
\end{table*}

\end{document}